\documentclass[letterpaper]{article} 
\usepackage{aaai2026}  
\usepackage{times}  
\usepackage{helvet}  
\usepackage{courier}  
\usepackage[hyphens]{url}  
\usepackage{graphicx} 
\usepackage{natbib}  
\usepackage{caption} 
\usepackage{algorithm}
\usepackage{algorithmic}

\usepackage{newfloat}
\usepackage{listings}

\usepackage{amsmath}
\usepackage{microtype}

\usepackage{inconsolata}

\usepackage{graphicx}
\usepackage{todonotes}
\usepackage{subcaption}
\usepackage{bm}
\usepackage{amsmath}
\usepackage{tikz}
\usepackage{booktabs} 
\usepackage{times}
\usepackage{latexsym}
\usepackage{multirow}
\usepackage{bm}
\usepackage{enumitem}
\usepackage{microtype}
\usepackage{amsfonts}
\usepackage{graphicx}
\usepackage{subcaption}
\usepackage[most]{tcolorbox}
\usepackage{color}
\usepackage{colortbl}
\usepackage{xcolor}         
\usepackage{nicefrac}
\definecolor{cblue}{rgb}{0.21,0.49,0.74}
\definecolor{pos}{rgb}{0.57,0.73,0.93}
\definecolor{neg}{rgb}{0.93,0.74,0.78}

\usepackage{pifont}
\usepackage{makecell}
\usepackage{mathtools}
\usepackage{fontawesome}
\usepackage[capitalize]{cleveref} 

\DeclareCaptionStyle{ruled}{labelfont=normalfont,labelsep=colon,strut=off} 
\floatstyle{ruled}
\newfloat{listing}{tb}{lst}{}
\floatname{listing}{Listing}
\title{What Makes a Good Speech Tokenizer for LLM-Centric Speech Generation? A Systematic Study}

\makeatletter
\let\AAAI@orig@copyright@text\copyright@text
\renewcommand{\copyright@text}{%
  \begingroup
    \footnotesize
    \def\@thefnmark{\faEnvelope}%
    \@makefntext{Corresponding authors are Tao Ji and Tao Gui.}%
    \par
  \endgroup
  \AAAI@orig@copyright@text
}
\makeatother

\author{
Xiaoran Fan\textsuperscript{1}\thanks{~Equal contribution.}, 
Zhichao Sun\textsuperscript{1}\footnotemark[1],
Yangfan Gao\textsuperscript{1}\footnotemark[1],
Jingfei Xiong\textsuperscript{1}\footnotemark[1], 
Hang Yan\textsuperscript{2}\footnotemark[1],
\\
\bf Yifei Cao\textsuperscript{1}, 
Jiajun Sun\textsuperscript{1},
Shuo Li\textsuperscript{1},
Zhihao Zhang\textsuperscript{1},
Zhiheng Xi\textsuperscript{1},
Yuhao Zhou\textsuperscript{1}, 
Senjie Jin\textsuperscript{1}, \\
\bf Changhao Jiang\textsuperscript{1}, 
Junjie Ye\textsuperscript{1}, 
Ming Zhang\textsuperscript{1}, 
Rui Zheng\textsuperscript{2}, 
Zhenhua Han\textsuperscript{2}, 
\\
Yunke Zhang\textsuperscript{3}, 
Demei Yan\textsuperscript{3}, 
Shaokang Dong\textsuperscript{3},
\bf Tao Ji\textsuperscript{1}\textsuperscript{\faEnvelope}, 
Tao Gui\textsuperscript{1,4}\textsuperscript{\faEnvelope}
}
\affiliations{
    \textsuperscript{1}Fudan University \\
\textsuperscript{2}Shanghai Qiji Zhifeng Co., Ltd \\
\textsuperscript{3}Honor Device Co., Ltd \\
\textsuperscript{4}Shanghai Innovation Institute \\
{{xrfan23@m.fudan.edu.cn; \{taoji,~tgui\}@fudan.edu.cn}
}
}

\begin{document}
\maketitle
\begin{abstract}
    Speech-language models (SLMs) offer a promising path toward unifying speech and text understanding and generation.
However, challenges remain in achieving effective cross-modal alignment and high-quality speech generation. 
In this work, we systematically investigate the role of speech tokenizer designs in LLM-centric SLMs, augmented by speech heads and speaker modeling.
We compare coupled, semi-decoupled, and fully decoupled speech tokenizers under a fair SLM framework and find that decoupled tokenization significantly improves alignment and synthesis quality.
To address the information density mismatch between speech and text, we introduce multi-token prediction (MTP) into SLMs, enabling each hidden state to decode multiple speech tokens.
This results in up to $12\times$ faster decoding and a substantial reduction in word error rate (from 6.07 to 3.01). 
Furthermore, we propose a speaker-aware generation paradigm and introduce RoleTriviaQA, a large-scale role-playing knowledge QA benchmark with diverse speaker identities. 
Experiments demonstrate that our methods enhance both knowledge understanding and speaker consistency.
\end{abstract}

\begin{links}
\link{Code}{https://github.com/cnxupupup/SLM-Decoupled-MTP}
\link{Demo}{https://cnxupupup.github.io/SLM-Decoupled-MTP-Demo}
\link{Dataset}{https://huggingface.co/datasets/cnxup/RoleTriviaQA}
\end{links}

\section{Introduction}
\label{sec:intro}
In recent years, large language models (LLMs) have achieved remarkable progress in text understanding and generation~\cite{grattafiori2024llama3herdmodels,qwen2025qwen25technicalreport}, fueling rapid advancements in multimodal models such as vision-language models~\cite{DBLP:conf/nips/LiuLWL23a}, vision-language-action models~\cite{DBLP:conf/corl/KimPKXB0RFSVKBT24}, and speech-language models (SLMs)~\cite{chu2023qwenaudioadvancinguniversalaudio,li2025baichuanaudiounifiedframeworkendtoend}. 
Among them, SLMs are particularly valuable for applications like natural and fluent human-computer dialogue~\cite{defossez2024moshispeechtextfoundationmodel,xu2025qwen25omnitechnicalreport} and personalized speech generation~\cite{du2024cosyvoicescalablemultilingualzeroshot,du2024cosyvoice2scalablestreaming}. 
The core challenge lies in generalizing the language understanding and generation capabilities of LLMs to incorporate and integrate speech understanding and generation.

Current approaches for connecting text and speech in SLMs can be broadly categorized into two perspectives.
\emph{Speech-centric} models~\cite{du2024cosyvoicescalablemultilingualzeroshot, du2024cosyvoice2scalablestreaming, anastassiou2024seedttsfamilyhighqualityversatile} are built upon mature speech generation architectures, with LLMs introduced as a conditional context. 
However, while this approach achieves high-quality speech synthesis, it struggles to fully leverage LLMs' rich world knowledge and powerful capabilities.
\emph{LLM-centric} models~\cite{defossez2024moshispeechtextfoundationmodel, DBLP:conf/emnlp/ZhangLZZWZQ23, DBLP:conf/acl/ZhanDYZZLZYZL0F24} take LLMs as the backbone and extend them with speech interaction. 
In theory, it fully utilizes LLMs' strengths (e.g., intent understanding, instruction following, planning, and decision-making). 
However, the quality of speech generation often suffers due to challenges in cross-modal alignment.
This work focuses on the \emph{LLM-centric} SLMs, aiming to explore how core components affect cross-modal alignment and the quality of generated speech.

The \textbf{speech tokenizer} and \textbf{speech head} determine the atomic inputs and latent representations of speech signals, which are critical to cross-modal alignment. 
Existing \textbf{speech tokenizers} can be categorized into three types: \emph{coupled} tokenizers (e.g., WavTokenizer (\citeyear{ji2025wavtokenizer})) that jointly encode semantic and acoustic details; \emph{semi-decoupled} tokenizers (e.g., SpeechTokenizer (\citeyear{DBLP:conf/iclr/ZhangZLZQ24})) that distill semantic content via HuBERT (\citeyear{DBLP:journals/taslp/HsuBTLSM21}) while retaining partial coupling; and \emph{decoupled} tokenizers (e.g., FACodec (\citeyear{DBLP:conf/icml/JuWS0XYLLST000024})) that separate speech into independent subspaces such as semantics, prosody, and timbre. 
Although all three achieve comparable performance in speech reconstruction tasks, their compatibility with SLMs remains underexplored. 
Under fair comparison within SLMs, we find that decoupled tokenizers are more favorable for cross-modal alignment and yield higher speech synthesis quality.

We observe a mismatch in information density between speech and text representations in the \textbf{speech head}: one second of speech corresponds to hundreds of tokens, far exceeding the amount of text information humans typically express per second. 
In LLMs, multi-token prediction (MTP)~\cite{DBLP:conf/icml/GloeckleIRLS24, DBLP:conf/icml/CaiLGPLCD24,deepseekai2025deepseekv3technicalreport} is often used to accelerate inference or enrich hidden representations by enabling a single hidden state to predict multiple future tokens. 
To balance the information density between speech and text, we introduce MTP into SLMs, allowing each speech hidden vector to decode multiple speech tokens (for both coupled and decoupled tokenizers).
We experimented with various compression ratios, ranging from one token per hidden state to twelve tokens per state. 
The MTP not only improved speech decoding speed by up to 12$\times$ but also significantly enhanced speech synthesis quality (the Word Error Rate decreased from 6.07 to 3.01).

In addition, we propose a speaker-aware speech generation paradigm that introduces speaker identity features into the speech-language context to guide and control the speaker's timbre in synthesized speech. 
To comprehensively evaluate SLMs’ understanding and generalization of the general knowledge captured by LLMs, as well as their speaker consistency, we construct a role-playing knowledge QA task. 
Specifically, we use CosyVoice2 (\citeyear{du2024cosyvoice2scalablestreaming}) to convert answers from a text-based knowledge QA dataset—TriviaQA (\citeyear{DBLP:conf/acl/JoshiCWZ17}) into speech, assigning each answer to one of 15 Genshin characters. 
The resulting RoleTriviaQA dataset contains 138K/0.3K/2.4K samples for training/validation/test, respectively. 
We reserve five characters and the WebQuestions (\citeyear{li2025baichuanaudiounifiedframeworkendtoend}) test set for out-of-domain evaluation of both speaker consistency and knowledge QA performance (i.e., evaluating generalization).
Results show that the decoupled architecture enables cooperative improvement in knowledge QA tasks and speaker timbre, and significantly outperforms other baselines.
\section{Speech-Language Model}
\label{sec:method}

\begin{figure*}[htp]
    \centering
    \includegraphics[width=0.9\textwidth]{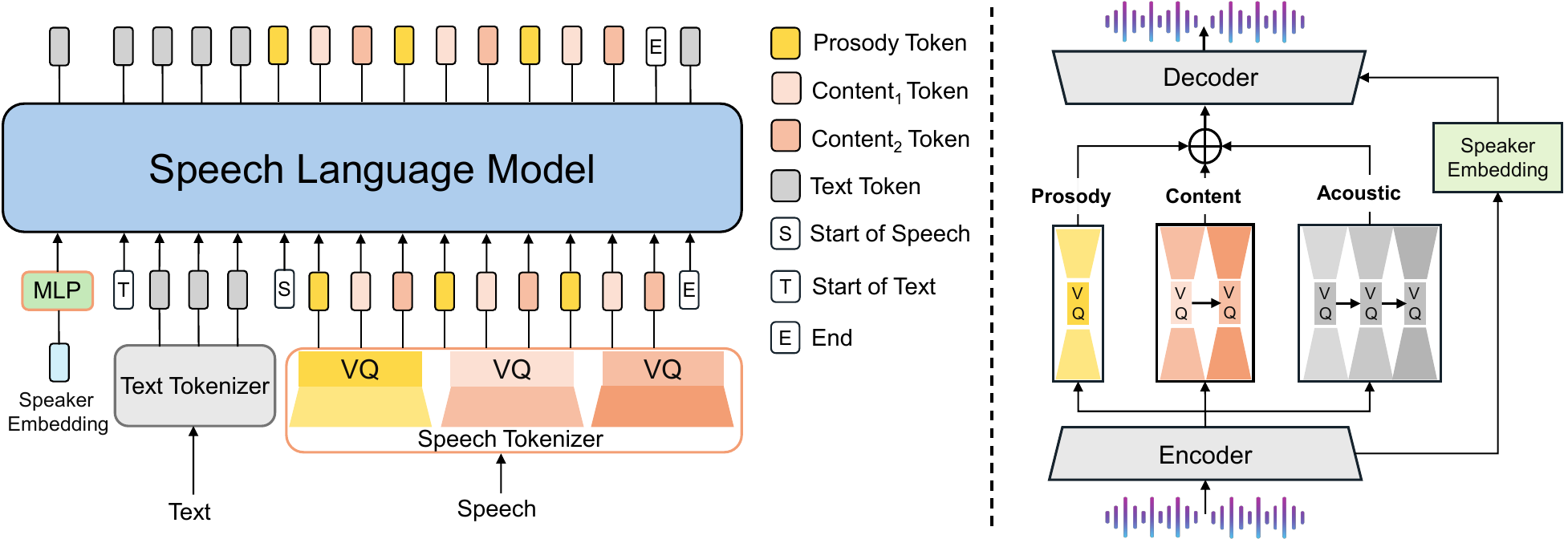}
    \caption{\textbf{Left}: Overview of a Speech Language Model (SLM) trained with a decoupled speech tokenizer (Section~\ref{ssec:slm}) and Speaker-Aware TTS (Section~\ref{ssec:sa_tts}); \textbf{Right}: The architecture of a possible decoupled speech tokenizer, featuring speech quantization, reconstruction in a decoupled manner, and speaker-specified embedding extraction.}
    \label{fig:llm}
\end{figure*}

A typical speech-language model (SLM) is usually an extension of a successful large language model (LLM) capable of understanding and generating both text and speech signals. 
It relies on a \emph{speech-language tokenizer} to convert raw text or audio into discrete tokens, a \emph{decoder-only Transformer} to learn cross-modal context and alignment, a \emph{language head} to predict text, and a \emph{speech head} to predict speech tokens and then synthesize continuous audio signals.

In the following, we introduce these core components and discuss how different design choices affect the SLM's capabilities, particularly in the speech-language tokenizer and speech head.

\subsection{Core Components}
\label{ssec:slm}

\paragraph{Speech-Language Tokenizer}
is a key component of SLMs, as it defines the atomic representations of text and speech. 
High-quality representations are crucial for effective cross-modal alignment, efficient generation, and preservation of acoustic details, making this component the focus of extensive research~\cite{DBLP:conf/acl/ZhanDYZZLZYZL0F24,DBLP:conf/acl/Peng00MH24}. 
It converts text or speech into discrete token sequences, such as $X_{\text{\faFileTextO}}=\{t_1, t_2, \dots, t_{|\text{\small\faFileTextO}|}\}$ for text and $X_{\text{\faFileSoundO}}\!=\!\{s_1, s_2, \dots s_{|\text{\small \faFileSoundO}|}\}$ for speech.
These sequences are ultimately combined into interleaved text-speech sequences $X\!=\!\{\dots, X_{\text{\faFileSoundO}}, X_{\text{\faFileTextO}}, X_{\text{\faFileSoundO}}, \dots\}$ to handle a variety of speech-language tasks, such as ASR~\cite{DBLP:conf/icassp/FathullahWLJSLG24}, TTS~\cite{du2024cosyvoicescalablemultilingualzeroshot,du2024cosyvoice2scalablestreaming}, speech translation~\cite{chen2024llastimprovedendtoendspeech}, and spoken question answering~\cite{zhao2024librisqanoveldatasetframework}.

Speech tokenizers can be categorized into coupled $X_{\text{\faFileSoundO}}\!=\!\{s_1, s_2, \dots\}$ and decoupled $X_{\text{\faFileSoundO}}\!=\!\{\bar{s}_1, \tilde{s}_1, \bar{s}_2, \tilde{s}_2, \dots\}$ 
\footnote{
    For simplicity, we assume a two-way decomposition, though in practice it can be readily extended to more components, e.g., $\hat{s}_i, \check{s}_i, \dots$.
}
\textsuperscript{,}
\footnote{
    Decoupled sequences may also be arranged in a type-wise contiguous manner within a chunk (e.g., $X_{\text{\faFileSoundO}} = \{\bar{s}_1, \bar{s}_2, \cdots, \tilde{s}_1, \tilde{s}_2, \dots\}$) rather than in an interleaved fashion. 
    However, experiments in ~\Cref{app:ntp_selection_form} 
    show that interleaved ordering yields significantly better performance.
}
designs for semantic $\bar{s}_*$ and acoustic details $\tilde{s}_*$. 
Intuitively, as illustrated in~\Cref{fig:llm}, decoupled semantic tokens are easier to align with text tokens, while the SLM can leverage its overparameterization to handle the acoustic detail tokens specifically. 
In contrast, coupled tokenizers may compromise cross-modal semantic alignment.

However, current comparisons between speech tokenizers are primarily conducted on speech reconstruction tasks, which are independent of SLMs. 
As a result, it is difficult to accurately assess their suitability for SLMs. 
For example, under fair conditions, the coupled BigCodec (\citeyear{xin2024bigcodecpushinglimitslowbitrate}) achieves better reconstruction quality, yet when integrated into an SLM, the decoupled FACodec (\citeyear{DBLP:conf/icml/JuWS0XYLLST000024}) yields significantly better performance.
This paper provides a detailed analysis of mainstream speech tokenizers within the context of SLM evaluation.

\paragraph{Decoder-only Transformer}
has achieved great success as a foundation model for text understanding and generation. 
After unifying the tokenization of text and speech, it can further learn joint speech-text context and perform cross-modal alignment. 
Its parameters are typically initialized from a well-pretrained LLM and then adapted to multimodal contexts.
Given a sequence of input tokens $X$, it produces a hidden representation vector $\bm{h}_i \in \mathbb{R}^d,~ H = \{\bm{h}_1, \bm{h}_2, \dots, \bm{h}_{|X|}\}$ and for each token, $   \mathbf{h}_i = \operatorname{Transformer}(X_{\le i}).$

\paragraph{Prediction Heads} include a \textit{language head} and a \textit{speech head}. 
The language head is consistent with that of an LLM and can be initialized from pretrained parameters. 
In contrast, the speech head has a different ``vocabulary size'' and is randomly initialized.
\textbf{Language head} consists of a linear transformation followed by a softmax function, 
\begin{align}
P\left(t_{i+1} \mid X_{\le i} \right) = \operatorname{softmax}_{~t_{i+1}} (W_{\text{\small\faFileTextO}} \cdot \bm{h}_i).
\end{align}

A parameter matrix $W_{\text{\small\faFileTextO}} \in \mathbb{R}^{|V_{\text{\small\faFileTextO}}| \times d}$ maps the hidden vector $\bm{h}_i$ to the vocabulary space of size $|V_{\text{\small\faFileTextO}}|$.
During training, the model is optimized using the cross-entropy loss: $\mathcal{L}_{\text{\scriptsize\faFileTextO}} = -\log P\left(t_{i+1} \mid X_{\le i} \right)$,
and during inference, the next token is predicted by either taking the $\arg\max P(t_{i+1} \mid X_{\le i})$ or sampling.
\textbf{Speech head} follows a similar formal definition, training objective, and inference process as the language head,
\
\begin{align} 
P\big({s}_{i+1} \mid X_{\le i}\big)= 
\operatorname{softmax}_{~{s}_{i+1}} (W_{\text{\small\faFileSoundO}} \cdot \bm{h}_i).
\label{eq:speech_head_coupled_ntp}
\end{align}
It uses a parameter matrix $W_{\text{\small\faFileSoundO}} \in \mathbb{R}^{|V_{\text{\small\faFileSoundO}}| \times d}$ to map the hidden vector $\bm{h}_i$ to the speech ``vocabulary'' of size $|V_{\text{\small\faFileSoundO}}|$.
When semantic and acoustic components are decoupled (i.e., $s_i$ is split into $\bar{s}_i$ and $\tilde{s}_i$), the vocabulary and the linear projection are also split accordingly into independent parts:

\begin{align} 
\left\{
\begin{aligned}
P\big(\bar{s}_{i+1} \mid X_{\le i}\big)&= 
\operatorname{softmax}_{~\bar{s}_{i+1}} (\overline{W}_{\text{\small\faFileSoundO}} \cdot \bm{h}_i)\\
P\big(\tilde{s}_{i+1} \mid X_{\le i}\big)&= 
\operatorname{softmax}_{~\tilde{s}_{i+1}} (\widetilde{W}_{\text{\small\faFileSoundO}} \cdot \bm{h}_i),
\end{aligned}
\right.
\label{eq:speech_head_decoupled_ntp}
\end{align}
where $ \overline{W}_{\text{\small\faFileSoundO}} \in \mathbb{R}^{|\overline{V}_{\text{\small\faFileSoundO}}| \times d}$ and $\widetilde{W}_{\text{\small\faFileSoundO}} \in \mathbb{R}^{|\widetilde{V}_{\text{\small\faFileSoundO}}| \times d}$.

Finally, the speech is reconstructed through the audio decoder from a pretrained codec: 
$
    \text{\faFileSoundO} = \operatorname{Decoder}(X_{\text{\faFileSoundO}}).
$

\subsection{Multi Token Prediction}
\label{ssec:mtp}
\begin{figure}[t]
    \centering
    \includegraphics[width=0.9\columnwidth]{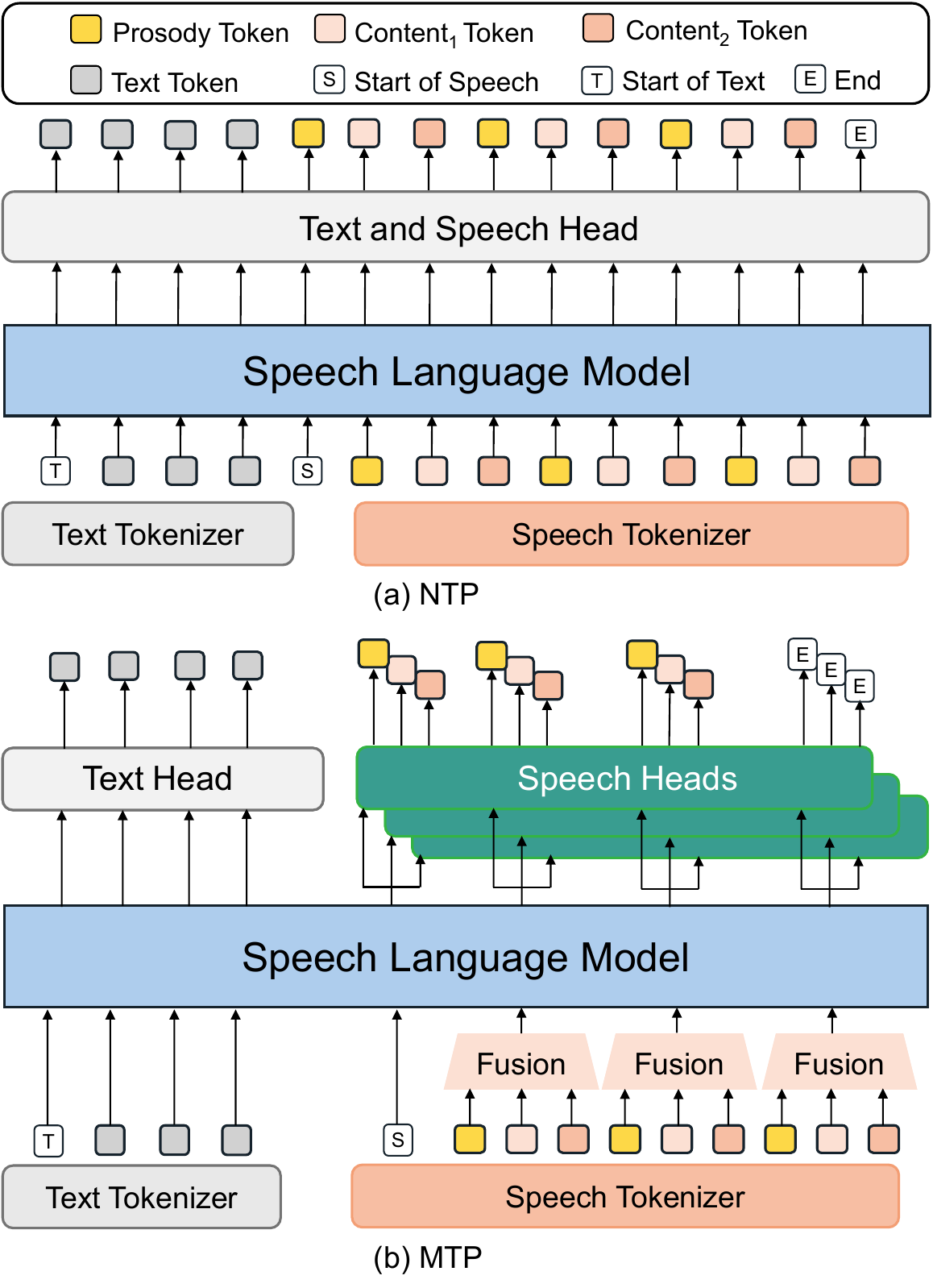}
    \caption{Illustration of our NTP and MTP architecture. (a) NTP: single vocabulary and single prediction head; (b) MTP: multiple vocabularies and multiple prediction heads, generating multiple tokens in parallel.}
    \label{fig:ntp_mtp}
\end{figure}

We observe that text representations are significantly more information-dense than speech. 
Typically, one second of speech is decoded into several hundred tokens, while it rarely conveys more than $\sim$20 text tokens' worth of information.
This asymmetry in information density can make cross-modal alignment difficult to learn.
On the other hand, speech tokens with excessively low information density can also hinder SLM inference efficiency, as generating one second of speech may take significantly longer than one second.
Therefore, balancing the information density between speech and text representations becomes a critical challenge.

Multi-token prediction (MTP)~\cite{DBLP:conf/icml/GloeckleIRLS24,DBLP:conf/icml/CaiLGPLCD24,deepseekai2025deepseekv3technicalreport} is commonly used in LLMs to accelerate inference or enrich hidden representations (e.g., by requiring $\bm{h}_i$ to decode not only $t_{i+1}$ but also future tokens like $t_{i+2}, \ldots$).
To compress the information density of speech representations, we introduce MTP into the SLM framework. As depicted in~\Cref{fig:ntp_mtp}, compared with Next Token Prediction (NTP), MTP enables each hidden vector $\bm{h}_i$ to decode multiple speech tokens, whether coupled or decoupled, thus alleviating the imbalance in information density between speech and text.

We group every $g$ adjacent speech tokens into a multi-token group $G_j$, which is decoded simultaneously at step $j$, meaning all tokens in $G_j$ are predicted from the same hidden vector $\bm{h}_j$.
This significantly increases the information density encoded in $\bm{h}_j$, making it more compatible with text tokens.
The definitions of the multi-token groups $G_{j+1}^{(coupled)}$/$G_{j+1}^{(decoupled)}$
\footnote{
Where the group $G_{j+1}^{(decoupled)}$ consists of half semantic tokens
$\bar{s}_*$ and half acoustic detail tokens $\tilde{s}_*$.
}
for coupled/decoupled tokenizers subject to $\left|G_{j+1}^{(*)}\right| = g$ are as follows:
\begin{align}
G_{j+1}^{(coupled)} &= \{s_{j\times g+1}, s_{j\times g+2}, \dots, s_{j\times g+g}\}  \nonumber\\ 
G_{j+1}^{(decoupled)} &= \{\bar{s}_{j\times \frac{g}{2}+1}, \tilde{s}_{j\times \frac{g}{2}+1}, \dots, \nonumber\\
&~~~~~~~\dots, \bar{s}_{j\times \frac{g}{2}+\frac{g}{2}}, \tilde{s}_{j\times \frac{g}{2}+\frac{g}{2}}\}.  \nonumber
\end{align}
The parameter matrix of the speech head is extended from a 2D tensor to a 3D tensor to enable simultaneous prediction of multiple tokens.
The extended dimension has size $g$, where each slice corresponds to a linear projection matrix (as in \Cref{eq:speech_head_coupled_ntp} for coupled or \Cref{eq:speech_head_decoupled_ntp} for decoupled tokens) for one token in the group $G$,
\begin{align*}
 \bm{W}^{(coupled)}_{\text{\small\faFileSoundO}} &= [W^1_{\text{\small\faFileSoundO}}, W^2_{\text{\small\faFileSoundO}},\dots, W^g_{\text{\small\faFileSoundO}}],\\
 \bm{W}^{(decoupled)}_{\text{\small\faFileSoundO}} &= [\overline{W}^1_{\text{\small\faFileSoundO}}, \widetilde{W}^1_{\text{\small\faFileSoundO}},\dots, \overline{W}^{\nicefrac{g}{2}}_{\text{\small\faFileSoundO}}, \widetilde{W}^{\nicefrac{g}{2}}_{\text{\small\faFileSoundO}}].
\end{align*}
Similarly, the softmax normalization is applied after the linear transformations to yield the probability distributions for all tokens in the group,
\begin{align*}
 P\big(G_{j+1}^{(*)} \mid X_{\le j\times g}\big) &=  \operatorname{softmax}_{G_{j+1}^{(*)}}(\bm{W}^{(*)}_{\text{\small\faFileSoundO}} \cdot \bm{h}_j).
\end{align*}
The loss $\mathcal{L}_{\text{\scriptsize\faFileSoundO}}^{(coupled)}$/$\mathcal{L}_{\text{\scriptsize\faFileSoundO}}^{(decoupled)}$ is also extended to the average over all tokens within the group:
$\mathcal{L}_{\text{\scriptsize\faFileSoundO}}^{(*)}=\frac{1}{g}\sum_{s_k \in G_{j+1}^{(*)}} -\log P(s_k \mid X_{\le j \times g}).$

Finally, the Transformer input must be adapted for MTP. 
For each group, we first obtain the embeddings of all tokens (i.e., $\bm{s}_j = \operatorname{E}(s_j)$ in NTP), concatenate them, and feed the result into a fusion network like a linear or MLP to produce a downsampled (by a factor of $\nicefrac{1}{g}$) input vector:
\begin{align}
    \bm{s}_j &= \underbrace{\operatorname{Fusion}}_{e.g., \operatorname{MLP}}\Big(\bigoplus_{s_k \in G_{j}^{(*)}} \operatorname{E}(s_k)\Big),
\end{align}
where the $\bigoplus$ is the concatenation function.

\subsection{Speaker-Aware Speech Generation}
\label{ssec:sa_tts}

\begin{figure}[htbp!]
    \centering
    \includegraphics[width=0.9\columnwidth]{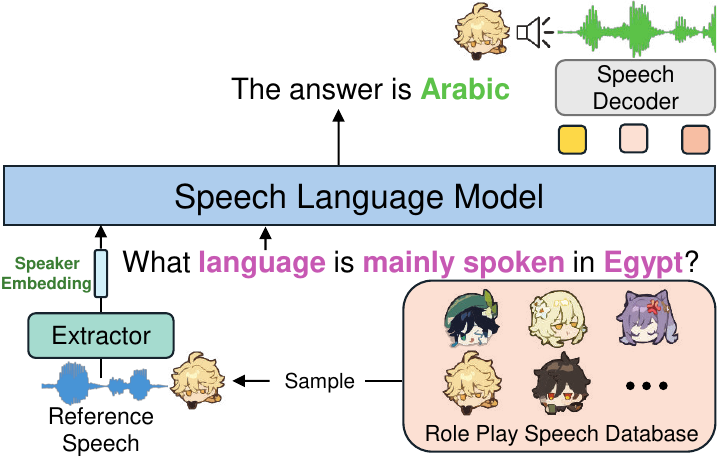}
    \caption{Illustration of Role-Playing Knowledge QA.
    }
    \label{fig:role_play}
\end{figure}

In speech generation (especially within spoken dialogue systems), explicitly controlling paralinguistic features such as speaker timbre, intonation, and emotional tone enables the generation of consistent and controllable speech responses.
Therefore, we introduce speaker-specific representations into the SLM context and evaluate both speech quality and character consistency in a role-playing speech QA task.

Inspired by~\citet{du2024cosyvoicescalablemultilingualzeroshot} and~\citet{li2024styletalkerfinetuningaudiolanguage}, we employ a pretrained timbre extractor
\footnote{
We adopt the timbre extractor from the decoupled tokenizer FACodec~\cite{DBLP:conf/icml/JuWS0XYLLST000024} in this work. 
Implementation details are provided in  \Cref{app:speaker_embedding}. 
It can be smoothly replaced with other extractors if needed.
}
to obtain the timbre representation $X_{\text{\faUser}}$ of the speaker or role. 
This representation can either be a discretized token  $X_{\text{\faUser}} = \{u_1, u_2, \dots, u_{|\text{\small\faUser}|}\}$, or a continuous vector sequence $X_{\text{\faUser}} = \{\bm{u}_1, \bm{u}_2, \dots, \bm{u}_{|\text{\small\faUser}|}\}$, where $|\text{\footnotesize\faUser}|$ is the sequence length.
The speech-language context of the SLM is then extended as follows: $ X = \{X_{\text{\faUser}}, \dots, X_{\text{\faFileSoundO}}, X_{\text{\faFileTextO}}, X_{\text{\faFileSoundO}}, \dots\}.
$

We expect the generated speech to be guided and controlled by the speaker-specific features.

\begin{table*}[htb]
    \centering
    \small
    \setlength{\tabcolsep}{2mm}
    \begin{tabular}{l|ccc|ccccc}
    \toprule
    \multirow{2}{*}{Model} & \multirow{2}{*}{\begin{tabular}[c]{@{}c@{}}Codebook \\ Size\end{tabular}} & \multirow{2}{*}{Nq} & \multirow{2}{*}{\begin{tabular}[c]{@{}c@{}}Token Rate \\ (TPS)\end{tabular}}  & \multirow{2}{*}{\begin{tabular}[c]{@{}c@{}} Success Rate\\ (SR) $\uparrow$\end{tabular}} & \multirow{2}{*}{UTMOS $\uparrow$} & \multirow{2}{*}{WER $\downarrow$} & \multicolumn{2}{c}{SIM $\uparrow$} \\ \cmidrule(lr){8-9}
& & & & & & & Seen & Unseen \\ \midrule
    \rowcolor{gray!10} \multicolumn{9}{c}{\textit{\textbf{Coupled Tokenizers trained with SLM}}} \\
    Encodec~\citeyearpar{DBLP:journals/tmlr/DefossezCSA23}         & 1024       & 4  & 300     & 0.38 & 2.86 & 17.00 & 0.18 & 0.10 \\
    Encodec~\citeyearpar{DBLP:journals/tmlr/DefossezCSA23}         & 1024       & 8  & 600     & 0.24 & 3.50 & 8.14 & 0.18 & 0.10 \\
    WavTokenizer~\citeyearpar{ji2025wavtokenizer}    & 4096       & 1  & 75      & 0.38 & 2.80 & 89.60 & 0.06 & 0.04 \\
    {WavTokenizer-v2}~\citeyearpar{ji2025wavtokenizer}    & 4096  & 1  & 75      & 0.65 & 3.16 & 49.24 & 0.11 & 0.08 \\
    StableCodec~\citeyearpar{DBLP:conf/iclr/ParkerSPCZEL25}     & 15625      & 2  & 50      & 0.98 & 3.90 & 13.04 & 0.17 & 0.14 \\
    BigCodec~\citeyearpar{xin2024bigcodecpushinglimitslowbitrate}        & 8192       & 1  & 80      & 0.70 & 3.93  & 13.63 & 0.18 & 0.10 \\

    \rowcolor{gray!10} \multicolumn{9}{c}{\textit{\textbf{Semi-Decoupled Tokenizers trained with SLM}}} \\
    SpeechTokenizer~\citeyearpar{DBLP:conf/iclr/ZhangZLZQ24} & 1024       & 5  & 250     & \underline{0.99} & 3.86 & 13.13 & 0.16 & 0.13 \\
    SpeechTokenizer~\citeyearpar{DBLP:conf/iclr/ZhangZLZQ24} & 1024       & 8  & 400     & 0.90 & \bf4.05 & 9.95 & 0.14 & 0.14 \\
    
    \rowcolor{gray!10} \multicolumn{9}{c}{\textit{\textbf{Decoupled Tokenizers trained with SLM}}} \\
    {FACodec-NTP-1H}         & 1024       & 3  & 240      & 0.87 & \underline{3.93} & 6.07 & \bf0.50 & \bf 0.49 \\

    \midrule
    {FACodec-MTP-3H}         & 1024       & 3  & 80    & 0.82 & 3.83 & 5.84 & \underline{0.47} & \underline{0.47} \\ 
    {FACodec-MTP-6H}         & 1024       & 3  & 40    & 0.92 & 3.75 & \underline{4.36} & \underline{0.47} & \underline{0.47} \\ 
    {FACodec-MTP-12H}         & 1024       & 3  & 20    & \bf1.00 & 3.67 & \bf 3.01 & \underline{0.47} & \underline{0.47} \\ 

    \bottomrule
    \end{tabular}
    \caption{Comparison of coupled, semi-decoupled, and decoupled tokenizers trained with SLMs (H: speech heads for FACodec-based SLMs). Evaluated on LibriTTS Test-Clean, with SIM tested on both seen (Train) and unseen (Test-Clean) speakers. Best and second-best results are highlighted in bold and \underline{underline}, respectively.
    Detailed information about these baselines can be found in Appendix~\ref{app:codec_compare_append}.}
    \label{tab:codec_compare}
\end{table*}

\begin{table}[t]
    \centering
    \small
    \setlength{\tabcolsep}{1.2mm} 
    \begin{tabular}{lccc|ccc}
    \toprule
    Model & H & TPS & Spk-Aware & UTMOS$\uparrow$ & WER$\downarrow$ & SIM$\uparrow$ \\ 
    \midrule
    {NTP}   & 1 & 240 & \ding{55} & \underline{3.93} & 6.07 & 0.49 \\
    {NTP}   & 1 & 240 & \ding{51} & \bf 4.05 & 5.51 & 0.57 \\ 
    \midrule
    {MTP}   & 3 & 80 & \ding{55} & 3.83 & 5.84 & 0.47 \\
    {MTP}   & 3 & 80 & \ding{51} & 3.90 & 7.04 & 0.56 \\ 
    {MTP}   & 6 & 40 & \ding{55} & 3.75 & 4.36 & 0.47 \\  
    {MTP}   & 6 & 40 & \ding{51} & 3.94 & 4.07 & \underline{0.59} \\ 
    {MTP}   & 12 & 20 & \ding{55} & 3.67 & \underline{3.01} & 0.47 \\ 
    {MTP}   & 12 & 20 & \ding{51} & \underline{3.93} & \bf 2.94 & \bf 0.60 \\ 
    \bottomrule
    \end{tabular}
    \caption{Impact of Speaker-Aware training for decoupled tokenizer with NTP and MTP architectures, evaluated on LibriTTS test-clean dataset.}
    \label{tab:spk_aware}
\end{table}

\section{Role-Playing Knowledge QA Task}

\label{ssec:role-play}
Current SLMs excel at daily conversations~\cite{defossez2024moshispeechtextfoundationmodel}, but their performance on knowledge-intensive tasks (e.g., QA) and voice characteristics (e.g., role-playing) lacks systematic evaluation. To address this problem, we introduce the \textbf{Role-Playing Knowledge QA} task, assessing SLM knowledge retention from its backbone LLM and similarity to reference speech in voice characteristics.

However, a critical gap exists: \textbf{no open-source dataset} assesses both correctness and timbre similarity in speech knowledge QA. Other models featuring speech QA~\cite{defossez2024moshispeechtextfoundationmodel} were trained and evaluated on a large-scale dataset, which has not been released yet.

To bridge this gap and stimulate SLMs' intrinsic knowledge with substantial training~\cite{DBLP:conf/emnlp/ZhangLZZWZQ23}, we propose \textbf{RoleTriviaQA}, an open-source dataset for benchmarking SLMs on role embodiment and knowledgeable spoken responses. Similar with~\cite{DBLP:conf/emnlp/ZhangLZZWZQ23}, we adopt the data format $X^{(role\text{-}qa)}\!=\! \{X_{\text{\small \faUser}}, X^{(q)}_{\text{\small\faFileTextO}}, X^{(a)}_{\text{\small\faFileTextO}}, X^{(a)}_{\text{\small\faFileSoundO}}\}$ (see~\Cref{fig:role_play}), which enables essential text-speech knowledge alignment for high-quality role-playing.

\subsection{Data Construction}
\label{ssec:qa_dataset_construction}

We created the RoleTriviaQA dataset by selecting 15 distinct voices from \textit{Genshin Impact}
roles (10 seen: 5 Male/5 Female for training and evaluation; 5 unseen: 2 Male/3 Female for evaluation only). Then we synthesize speech for TriviaQA~\citep{DBLP:conf/acl/JoshiCWZ17} QA-pairs using CosyVoice 2~\citep{du2024cosyvoice2scalablestreaming} after filtering high-quality reference speeches. This pipeline yields 138,384 training samples and a test set with 1,500 in-domain (ID, from TriviaQA) and 926 out-of-domain (OOD, from OpenAudioBench~\cite{li2025baichuanaudiounifiedframeworkendtoend}) samples. Further details of RoleTriviaQA are in~\Cref{app:role_play_dataset}.

\subsection{Training Strategy}
\label{ssec:qa_model_train}

To build an SLM support role-playing knowledge QA task, we adopt a two-stage training method:

\paragraph{Stage 1: Cross-Modal Alignment Pretraining}
To improve cross-modal alignment, we conduct a large-scale dataset \textbf{Emilia-3.5}.  
First, we filter Emilia (\citeyear{DBLP:conf/slt/HeSWLGHLYLSWCZW24}) (a large-scale, real-world speech collection with diverse emotions and styles) for samples scoring at least 3.5 on DNSMOS (\citeyear{DBLP:conf/icassp/ReddyGC22}). 
Next, we merge the standard LibriTTS (\citeyear{DBLP:conf/interspeech/ZenDCZWJCW19}) training set.
The resulting corpus comprises 2,495 hours of high-quality speech.
Moreover, to enable bidirectional alignment of text and speech modalities, we conduct pre-training on both TTS and ASR tasks.
Specifically, we adopt $X^{(tts)}\!=\!\{ X_{\text{\small \faUser}}, X_{\text{\small \faFileTextO}},X_{\text{\small \faFileSoundO}}\}$ as the TTS data format and $X^{(asr)}\!=\! \{X_{\text{\small \faFileSoundO}}, X_{\text{\small \faFileTextO}}\}$ for ASR. The pretaining loss $\mathcal{L}^{(pt)}$ is formulated as:
$\mathcal{L}^{(pt)}\!=\!\sum_{y_i \in X^{(tts/asr)}}\!-\!\log P(y_i \mid X^{(tts/asr)}_{<i}),
$
where the predicted token $y_i$ could be a single token or a multi-token group (i.e., $G_i$).

\paragraph{Stage 2: Knowledge-Role Joint Fine-Tuning}After the pretraining stage, we perform supervised fine-tuning (SFT) with RoleTriviaQA to enable the SLM to capture the voice features of each role along with the knowledge-intensive data.

The SFT loss $\mathcal{L}^{(role\text{-}qa)}$ is defined as follows:$\mathcal{L}^{(role\text{-}qa)}\!=\!\sum_{y_i\in X^{(a)}_{\text{\faFileTextO}\cup\text{\faFileSoundO}}}\!-\!\log P(y_i \mid  X^{(role\text{-}qa)}_{<i}),$
where the loss is calculated across both the text answer and the speech answer.

\section{Experiment}
\label{sec:exper}

\begin{table*}[htbp]    
    \centering
    \fontsize{8}{10}\selectfont
    \setlength{\tabcolsep}{1.5mm}
    \begin{tabular}{lc c ccccc ccccc}
    \toprule
    \multirow{3}{*}{\textbf{Models}} & 
    \multirow{3}{*}{\makecell{\textbf{Model} \\ \textbf{Type}}} & 
    \multirow{3}{*}{\makecell{\textbf{Tokenizer} \\ \textbf{Type}}}  
    & \multicolumn{5}{c}{\textbf{In-Domain (RoleTriviaQA)}} & \multicolumn{5}{c}{\textbf{Out-of-Domain (Web Questions)}} \\
    \cmidrule(lr){4-8} \cmidrule(l){9-13}
    & & & \multirow{2}{*}{EM(\%)$\uparrow$} & \multirow{2}{*}{F1(\%)$\uparrow$} & \multicolumn{3}{c}{SIM$\uparrow$} & \multirow{2}{*}{EM$\uparrow$} & \multirow{2}{*}{F1$\uparrow$} & \multicolumn{3}{c}{SIM$\uparrow$} \\
    & & & & & Seen & Unseen & All & & & Seen & Unseen & All \\
    \midrule
    Qwen2.5-0.5B-Instruct & LLM & -       & 7.1 & 17.3 & - & - & - & 1.1  & 14.0  & - & - & - \\
    \midrule
    WavTokenizer-v2       & SLM & Coupled & 5.1 & 12.5 & 0.22 & \underline{0.18} & 0.20 & \underline{1.1} & 3.9 & 0.21 & \underline{0.18} & 0.20 \\
    BigCodec              & SLM & Coupled & 2.9 & 9.4 & \underline{0.23} & 0.17 & \underline{0.21} & 1.0 & \underline{4.2} & \underline{0.23} & 0.17 & \underline{0.21} \\
    StableCodec           & SLM & Coupled      & \underline{7.5} & \underline{13.8} & 0.18 & 0.14 & 0.17 & 0.6 & 3.6 & 0.17 & 0.14 & 0.16 \\
    SpeechTokenizer       & SLM & Semi-Decoupled & 3.4 & 9.1 & 0.19 & 0.14 & 0.17 & 0.3 & 2.6 & 0.20 & 0.14 & 0.18 \\
    FACodec               & SLM & Decoupled      & \bf12.0 & \bf23.8 & \bf0.63 & \bf0.58 & \bf0.61 & \bf9.0 & \bf17.9 &\bf0.63 & \bf0.58 & \bf0.61 \\
    \bottomrule
    \end{tabular}
    \caption{Comparisons of coupled, semi-decoupled and decoupled tokenizers trained with SLMs on RoleTriviaQA test set.}
    \label{tab:role_play}
\end{table*}

Our experiments address two critical questions:  
\begin{enumerate}[leftmargin=*,itemsep=0pt, topsep=0pt, parsep=0pt]
    \item How do \textbf{different speech tokenizer} and \textbf{training methods} (NTP/MTP/Speaker-Aware) affect SLMs' performance? (Sections~\ref{ssec:res_different_tokenizer_slm}-\ref{ssec:res_speaker_aware_tts})
    \item How can SLMs effectively leverage their intrinsic knowledge while maintaining high-quality speech synthesis? (Section~\ref{ssec:res_role_play_knowledge_qa})
\end{enumerate}

\paragraph{Setups}
\label{ssec:exp-setup}
Details of the datasets, baselines, evaluation, and hyperparameter setups are placed in \Cref{app:settings}.

\subsection{Effect of Speech Tokenizers}
\label{ssec:res_different_tokenizer_slm}

As shown in \Cref{tab:codec_compare}, the degree of speech representation decoupling shows a significant correlation with the joint SLM training effect, where the semi-decoupled and fully decoupled architectures exhibit better overall performance.

First, coupled architectures exhibit severe convergence challenges during SLM training (see
Appendix C), resulting in generally low success rates of speech synthesis (\(\leq\)70\%). 
The notable exception is StableCodec, whose large codebook and low token rate establish dual compensation mechanisms that mitigate inherent limitations of coupled architectures.
Second, the semi-decoupled and decoupled architectures show clear advantages. The semi-decoupled architecture achieves the best speech quality, while the decoupled architecture works optimally in WER and SIM.
Third, the decoupled architectures’ SIM between the training and test sets is consistently larger and significantly better than the other baselines, suggesting superior generalization capabilities, especially for speaker timbre.

Overall, decoupled tokenizer and SLM joint modeling is better and has better controllability and scalability.

\subsection{Next Token vs. Multi-Token Prediction}
\label{ssec:res_ntp_mtp}
As illustrated in ~\Cref{tab:codec_compare}, to investigate the effectiveness of MTP under decoupled architecture, we also conduct experiments comparing models trained with different compression ratios.

First, the MTP-3H model is competitive with the NTP one, even with \(3\times\) compression. Moreover, it has also surpassed all of the semi-decoupled and coupled baselines in WER and SIM.
Second, when increasing compression from \(3\times\) (MTP-3H) to \(12\times\) (MTP-12H), WER improves by 48\% (5.84 to 3.01), with stable SIM and UTMOS experiencing a slight decline. 
Notably, under $12\times$ compression, our model achieves speech synthesis with a 100\% success rate, and the WER (3.01) is comparable to the ground truth. 
These results reveal that speech length compression based on the decoupled tokenizer effectively mitigates the frequency mismatch between text ($\sim$20Hz) and speech (240Hz) modalities, enhancing cross-modal alignment performance.

\subsection{Speaker-Aware Speech Generation}
\label{ssec:res_speaker_aware_tts}
\cref{tab:spk_aware} presents that Speaker-Aware training exhibits consistent enhancement effects across NTP and MTP architectures with a decoupled tokenizer.

First, for NTP, the incorporation of speaker embedding leads to consistent improvements across all metrics. 
Second, for MTP, speaker-embedding provides a stable gain in most cases.
In addition, as compression increases from 3H to 12H, Speaker-Aware models also show continuous WER reduction (7.04 to 2.94), 
In particular, the advantages of Speaker-Aware training become more evident at higher compression. For instance, the MTP-12H trained with speaker embedding significantly outperforms its non-aware model.

This indicates that incorporating speaker embeddings enables the model better to capture the speaker's paralinguistic features and linguistic accuracy.

\subsection{Role-Playing Knowledge QA}
\label{ssec:res_role_play_knowledge_qa}

\paragraph{Settings} 
We evaluate Role-Playing Knowledge Speech QA within two aspects: 1) Benchmark of the decoupled SLM against top-3 coupled/semi-decoupled SLMs (From~\Cref{tab:codec_compare}) and the LLM backbone (Qwen2.5-0.5B-Instruct); 2) Compare the decoupled SLM against diverse SLM baselines.

\paragraph{Results on Different Tokenizers} 
\Cref{tab:role_play} shows that SLM trained with a decoupled tokenizer achieves SOTA performance across all metrics. For ID evaluation, it outperforms all baselines with 12.0 EM and 23.8 F1, while its 0.61 SIM score is $2.9\times$ higher than the best coupled baseline (BigCodec). Though all models degrade on OOD queries, the decoupled model shows greater robustness, confirming that decoupling resolves target conflicts in semi-decoupled/coupled systems by jointly optimizing semantic and acoustic objectives.

Compared to the LLM baseline, the decoupled SLM significantly outperforms in both ID/OOD settings (OOD EM: 9.0 vs 1.1; F1: 17.9 vs 14.0), effectively preserving LLM knowledge while adapting to unseen queries. It also generalizes well in speaker timbre, achieving high SIM scores for both seen/unseen speakers (0.63 vs 0.58) with minimal gaps.

Overall, the decoupled architecture thus enables cooperative improvement in knowledge QA and speaker timbre, surpassing all baselines.

\paragraph{Results on Different SLMs}
Comparisons with other SLM baselines also confirm that decoupled tokenizers help not only improve the speech quality but also preserve LLM knowledge. Detailed results are provided in~\Cref{app:role_play_slm}.

\begin{table*}[htbp!]
    \centering
    \small    
    \setlength{\tabcolsep}{2mm} 
    \begin{tabular}{lll|ccc|ccc|c}
    \toprule
    \multirow{2}{*}{\textbf{Models}} & \multirow{2}{*}{\textbf{Heads}} & \multirow{2}{*}{\textbf{TPS}} 
    & \multicolumn{3}{c|}{\textbf{Cosine Similarity}} & \multicolumn{3}{c|}{\textbf{Euclidean Distance}} & \multirow{2}{*}{\textbf{\makecell{Riemannian\\ Distance$\downarrow$}}} \\
    \cmidrule(lr){4-6}
    \cmidrule(lr){7-9}
    & & & TTSim & SSSim & STSim$\uparrow$ & TTDist & SSDist & STDist$\downarrow$ & \\ 
    \midrule
    NTP & 1 & 240 & 0.896 & 0.999 & 0.480 & 0.438 & 0.030 & 1.014 & 365.26 \\
    \midrule
    MTP & 3 & 80 & 0.986 & 0.999 & 0.967 & 0.147 & 0.052 & 0.244 & 288.39 \\
    MTP & 6 & 40 & 0.991 & 0.998 & \underline{0.981} & 0.116 & 0.055 & \underline{0.186} & \underline{268.19} \\  
    MTP & 12 & 20 & 0.996 & 0.998 & \bf0.986 & 0.078 & 0.063 & \bf0.158 & \bf 237.55 \\ 
    \bottomrule
    \end{tabular}
    \caption{Cross-modal alignment metrics (calculated on the hidden states of the last layer) across different speech token compression rates. \textbf{Metrics}: TTSim/SSSim (intra-modal cosine similarity: text vs. speech); STSim (cross-modal cosine similarity); TTDist/SSDist (intra-modal Euclidean distance: text vs. speech); STDist (cross-modal Euclidean distance).}\label{tab:speech_text_sim_dis}
\end{table*}

\subsection{Ablation Study}
\paragraph{MTP Architecture Selection} 
We analyze 6 different fusion strategies and speech head architectures to select the optimal MTP configuration (\textbf{MLP-Linear} combination), as elaborated in \Cref{app:mtp_selection}.
\paragraph{NTP Layer and Token Organization Selection}
We explore the best number of decoupled tokenizer codebook layers and speech token organization patterns for SLM training. Results show that \textbf{3 codebook layers} with \textbf{Frame-Wise Interleaving organization pattern } is the best-performing setting. Details are provided in \Cref{app:ntp_selection_layer} and~\Cref{app:ntp_selection_form}.
\paragraph{RoleTriviaQA Data Format Selection} We conducted experiments on 3 different data formats for training with RoleTriviaQA and found that the current data format we adopted (see~\Cref{ssec:role-play}) performs best across all baselines compared with others. This experiment is detailed in~\Cref{app:data_format_ablation}.
\subsection{Effect of MTP on Cross-modal Alignment}
\label{ssec:analysis}
\Cref{tab:codec_compare} and \Cref{tab:spk_aware} show that higher speech token compression in the MTP architecture consistently improves performance, especially in WER. To investigate the hypothesis that higher compression improves modal alignment, we conduct detailed quantitative 
and qualitative analyses.

\paragraph{Quantitative Analysis}
As shown in Table~\ref{tab:speech_text_sim_dis}, token compression improves cross-modal alignment and maintains intra-modal integrity, with MTP models outperforming NTP baselines.
First, when compression increases from $3\times$ to $12\times$, intra-modal similarities (TTSim, SSSim) remain high in all settings, while cross-modal similarity (STSim) increases steadily from 0.967 to 0.986. 
Second, the cross-model distance (STDist) decreases significantly from 0.244 to 0.158
, while intra-modal distances (TTDist, SSDist) stay stable. These trends align with WER improvements, suggesting that stronger compression encourages more isomorphic and effective speech-text representations.
Additionally, compared to NTP, all MTP settings exhibit better cross-modal alignment. This demonstrates that MTP not only maintains intra-modal coherence but also enhances cross-modal correspondence under token compression.

We also use Riemannian distance (\citeyear{bonnabel-sepulchre-2009-riemannian}) to quantify speech-text alignment (See~\Cref{app:metrics_qa} for details). \Cref{tab:speech_text_sim_dis} shows that with higher compression, the Riemannian distance decreases (288.39 to 237.55), indicating that aggressive compression not only reduces embedding distance but also better preserves their intrinsic geometric structure.
\paragraph{Qualitative Analysis}
We include complementary qualitative results in \Cref{app:umap_quantitative}.

\subsection{Effect of Tokenizer-MTP Combination}
To examine the effectiveness of different tokenizer coupling strategies in MTP SLMs, we perform comprehensive experiments in \Cref{app:mtp_role}.

\section{Related Work}
\label{sec:related-work}

\paragraph{Speech Tokenization} has advanced through RVQ-based architectures and quantization innovations: SoundStream (\citeyear{DBLP:journals/taslp/ZeghidourLOST22}) introduced the RVQ-VAE framework, enhanced by Encodec (\citeyear{DBLP:journals/tmlr/DefossezCSA23}, LSTM temporal modeling) and SpeechTokenizer (\citeyear{DBLP:conf/iclr/ZhangZLZQ24}, HuBERT semantic distillation). Recent ultra-low-bitrate methods include Improved RVQGAN (DAC,~\citeyear{DBLP:conf/nips/KumarSLKK23}) with quantizer dropout, WavTokenizer (\citeyear{ji2025wavtokenizer}, 0.9kbps via single-layer RVQ), BigCodec (\citeyear{xin2024bigcodecpushinglimitslowbitrate}, 1.04kbps via Convolution-LSTM projection), and StableCodec (\citeyear{DBLP:conf/iclr/ParkerSPCZEL25}, 0.7kbps using Transformer with FSQ (\citeyear{DBLP:conf/iclr/MentzerMAT24})). 

\paragraph{Speech Language Models}
LLMs are increasingly adopted for unified speech processing through two primary paradigms. The first integrates speech encoders with LLMs for comprehension tasks like ASR~\cite{ma2024embarrassinglysimpleapproachllm} and spoken language understanding~\cite{chu2023qwenaudioadvancinguniversalaudio}, achieving strong analytical capabilities but lacking generative functionality. The second targets speech generation via dual strategies: codec-based models (e.g., SpeechGPT,~\citeyear{DBLP:conf/emnlp/ZhangLZZWZQ23}) discretize speech into compressed tokens, while diffusion-enhanced approaches (e.g., CosyVoice,~\citeyear{du2024cosyvoicescalablemultilingualzeroshot,du2024cosyvoicescalablemultilingualzeroshot}) synthesize waveforms through LLM-diffusion. 

\paragraph{Multi Token Prediction}
has emerged as a critical technique in LLMs to accelerate inference. 
Detailed related works are placed on \Cref{app:related_word_mtp}.

\section{Conclusion}

This work investigates how LLM-centric SLMs can better align and generalize speech-text understanding and generation by revisiting core design choices.
We find that fully decoupled tokenizers and multi-token prediction significantly improve alignment efficiency and speech quality. 
Additionally, incorporating speaker-aware modeling and role-based QA evaluation reveals improved generalization and character consistency.
Our findings provide practical insights for building more capable and scalable SLMs.

\section*{Acknowledgements}
The authors wish to thank the anonymous reviewers for their helpful comments. This work was partially funded by National Natural Science Foundation of China (No.62476061, 62376061,62576106, 62506079), Shanghai Rising-Star Program (23QA1400200), and Natural Science Foundation of Shanghai (23ZR1403500). The computations in this research were performed using the CFFF platform of Fudan University.
\bibliography{aaai2026}

\appendix
\clearpage
\renewcommand{\thesection}{\Alph{section}}
\renewcommand{\thesubsection}{\Alph{section}.\arabic{subsection}}

\begin{table*}[htb]
\centering
\small
\setlength{\tabcolsep}{2mm}
\begin{tabular}{l|cccc|cccc}
\toprule
Model           & \begin{tabular}[c]{@{}c@{}}Codebook \\ Size\end{tabular} & Nq & \begin{tabular}[c]{@{}c@{}}Token Rate\\ (TPS)\end{tabular} & \begin{tabular}[c]{@{}c@{}}Bandwidth\\ (bps)\end{tabular} & STOI$\uparrow$ & \begin{tabular}[c]{@{}c@{}}PESQ$\uparrow$\end{tabular} & UTMOS$\uparrow$ & SIM$\uparrow$  \\ \midrule
Ground Truth    & -          & -  & -        & -          & 1.00 & 4.55          & 4.09     & 1.00 \\
\midrule
Encodec         & 1024       & 8  & 600     & 6000      & \underline{0.94} & 3.18       & 3.09  & \bf0.88 \\
WavTokenizer    & 4096       & 1  & 75      & 900       & 0.90 & 2.63       & 3.79  & 0.65 \\
StableCodec     & 15625      & 2  & 50      & 697       & 0.91 & 2.91       & \textbf{4.23}  & 0.62 \\
BigCodec        & 8192       & 1  & 80      & 1040      & 0.93 & \underline{3.27}       & \underline{4.11}  & 0.84 \\
SpeechTokenizer & 1024       & 8  & 400     & 1000      & 0.92 & 2.61       & 3.87  & 0.83 \\
FACodec     & 1024      & 6  & 480      &   4800     & \bf0.95 & \bf3.47       & 3.90  & \underline{0.87} \\
\bottomrule
\end{tabular}
\caption{Comparisons of various codec models for speech reconstruction on the LibriTTS test-clean dataset. Detailed information about these models can be found in Appendix~\ref{app:codec_compare_append}.}
\label{tab:codec_vq}
\end{table*}

\section{Details of Baselines}
\subsection{Compared Baselines}
\label{app:codec_compare_append}

\begin{itemize}
    \item \textbf{Encodec} \cite{DBLP:journals/tmlr/DefossezCSA23}: An RVQ-based codec designed for universal audio compression.\footnote{\scriptsize{\url{https://huggingface.co/facebook/encodec_24khz}}}
    \item \textbf{WavTokenizer} \cite{ji2025wavtokenizer}: A single VQ codebook-based tokenizer for universal audio.\footnote{\scriptsize{\url{https://huggingface.co/novateur/WavTokenizer-large-speech-75token}}}
    \item \textbf{StableCodec} \cite{DBLP:conf/iclr/ParkerSPCZEL25}: A residual FSQ-based tokenizer for speech.\footnote{\scriptsize{\url{https://huggingface.co/stabilityai/stable-codec-speech-16k}}}
    \item \textbf{BigCodec} \cite{xin2024bigcodecpushinglimitslowbitrate}: A VQ-based single-stream codec for speech.\footnote{\scriptsize{\url{https://huggingface.co/Alethia/BigCodec/resolve/main/bigcodec.pt}}}
    \item \textbf{SpeechTokenizer} \cite{DBLP:conf/iclr/ZhangZLZQ24}: An RVQ-based codec with semantic distillation for speech.\footnote{\scriptsize{\url{https://huggingface.co/fnlp/AnyGPT-speech-modules/tree/main/speechtokenizer}}}
\end{itemize}

\subsection{Results of Codec Reconstruction}
\begin{figure}[htb]
    \centering
    \includegraphics[width=0.95\columnwidth]{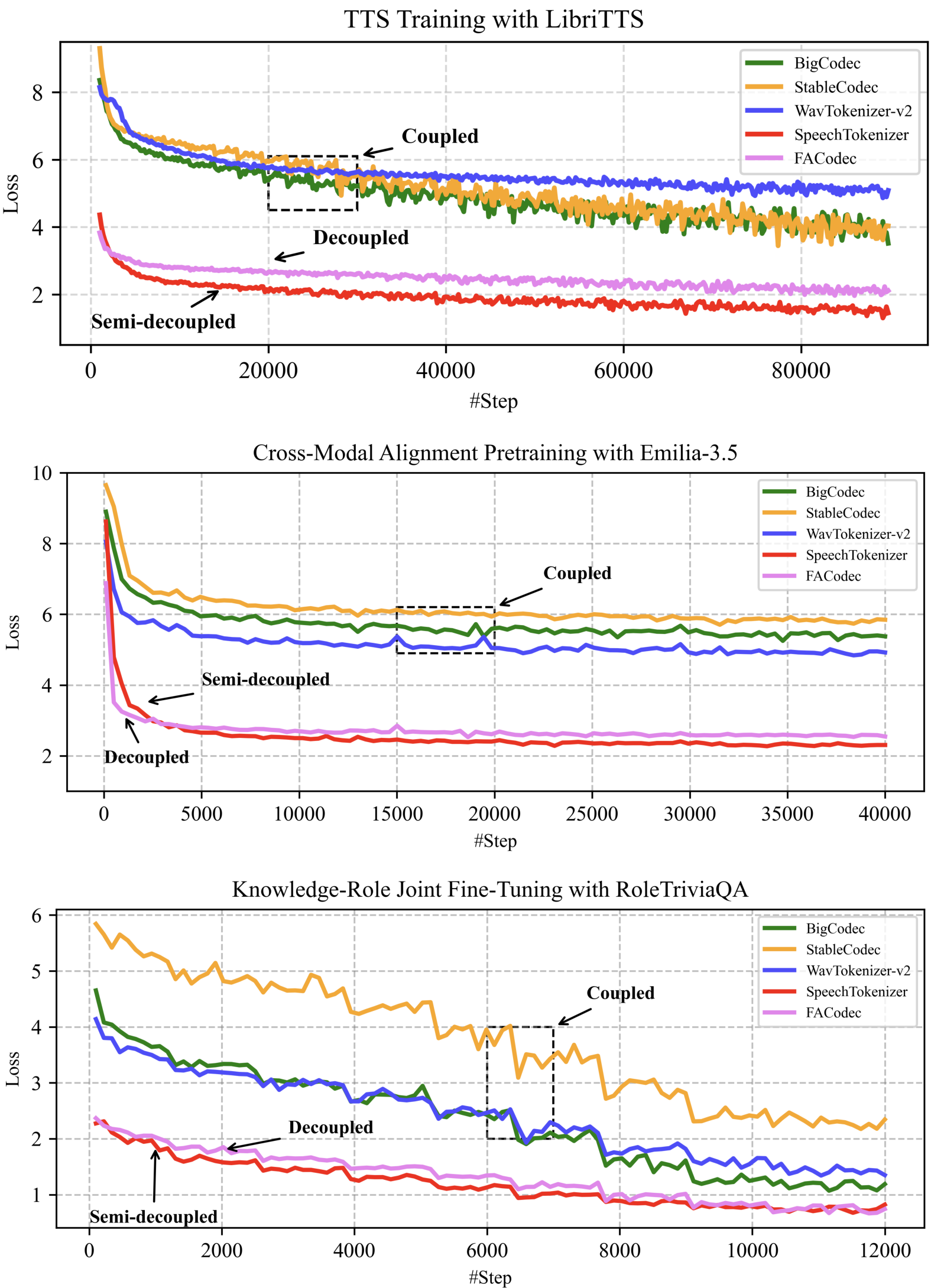}
    \caption{Training loss comparison between SLMs trained with baselines and FACodec across TTS training (upper figure) and 2-stage training for Role-Playing Knowledge QA task (middle and lower figures).}
    \label{fig:train-loss}
\end{figure}
\label{app:codec_vq_append}

Table~\ref{tab:codec_vq} shows the results of VQ reconstruction evaluation on the LibriTTS test-clean dataset.
The results reveal that for acoustic quality (STOI~\cite{DBLP:conf/icassp/TaalHHJ10}, PESQ~\cite{DBLP:conf/icassp/RixBHH01} and UTMOS), all tokenizers perform similarly with the ground truth, sharing close metric values. But for SIM, the trend differs: the VQ performs better on SIM as its token rate increases, and the decoupled model (FACodec) surpasses most of the coupled and semi-decoupled models. 
This is probably because reduced token rates compromise timbre fidelity, combined with coupled modeling diminishing acoustic prominence relative to semantic components.

\section{Speaker Embedding Extraction}
\label{app:speaker_embedding}
In FACodec, speaker information is extracted by a timbre extractor, which consists of four transformer encoder blocks. The module operates on the output $h$ of the speech encoder and transforms it into a global vector $X_{\text{\small \faUser}}$, which captures the characteristics of speakers. To generate audio, we sum the representations of prosody and content, and then integrate the speaker embedding via conditional layer normalization, enabling effective conditioning on speaker timbre during generation.

\section{Experimental Setups}
\label{app:settings}
\paragraph{Datasets}
For the default TTS training, we utilized the LibriTTS dataset, which contains 585 hours of English speech data. All audio samples are resampled to 16 kHz.

For Role-Playing Knowledge QA, we utilized the proposed Emilia-3.5 (2,495 hours) and RoleTriviaQA (138k) datasets during stages 1 and 2. Our proposed data details can be found in ~\Cref{ssec:role-play}.

\paragraph{Baselines}
We selected the current state-of-the-art speech tokenizer as the baseline for our model. To ensure a fair comparison, we employed the official weight files provided by the Encodec~\cite{DBLP:journals/tmlr/DefossezCSA23}, SpeechTokenizer~\cite{DBLP:conf/iclr/ZhangZLZQ24}, WavTokenizer~\cite{ji2025wavtokenizer}, StableCodec~\cite{DBLP:conf/iclr/ParkerSPCZEL25}, BigCodec~\cite{xin2024bigcodecpushinglimitslowbitrate}. Details can be found in \Cref{app:codec_compare_append}.

\paragraph{Training and Inference Hyperparameters}
Our baseline training configuration employs the AdamW optimizer ($\beta_1 = 0.9$ and $\beta_2 = 0.99$) with an initial learning rate of 5e-4 following a cosine decay schedule. All models undergo 90k training steps using consistent batch sizes of 16 for fair comparison. For specialized tasks like Role-Playing Knowledge QA, we implement a two-phase adaptation: 40k steps for cross-modal alignment pretraining followed by 12k steps of knowledge-role joint fine-tuning.
During inference, we apply greedy decoding with a 1.2 repetition penalty to balance output quality and diversity.
\paragraph{Hardware and Software Requirements} All experiments are conducted with 8*NVIDIA A800 GPUs and CUDA version 12.4 on Ubuntu 20.04.6 LTS.

\paragraph{Training Loss Comparison}
\Cref{fig:train-loss} depicts the training loss of NTP TTS training with LibriTTS and 2-stage training of the Role-Playing Knowledge QA task of all coupled, semi-decoupled baselines and FACodec, on which tokenizer types are annotated. It can be clearly seen that decoupled and semi-decoupled SLMs perform similarly and significantly better than coupled baselines across all tasks and stages. 

\paragraph{Evaluation Setups}
For the default TTS evaluation, following ~\cite{DBLP:conf/iclr/ZhangZLZQ24}, we randomly sample 300 samples from the LibriTTS test set for speech synthesis evaluation.
For Role-Playing Knowledge QA, the evaluation set of RoleTriviaQA contains 1,500 in-domain and 926 out-of-domain samples.

\paragraph{Metrics}
For the default TTS evaluation, we evaluate Success Rate (SR), speech quality (UTMOS~\cite{DBLP:conf/interspeech/SaekiXNKTS22}), Word Error Rate (WER), and speaker similarity (SIM). 

For the Role-Playing Knowledge QA task, we report three metrics: Exact Match (EM) and F1 score for text responses, while SIM for speech responses. Details of these metrics can be found in~\Cref{app:metrics_qa}.

\section{Details of Evaluation Metrics}
\begin{table}[htb]
    \centering
    \small
    \setlength{\tabcolsep}{0.8mm} 
    \begin{tabular}{l|cc|cc}
        \toprule
        Model & Model Size & Data Format & EM$\uparrow$ & F1$\uparrow$ \\
        \midrule
        SpiritLM(\citeyear{nguyen2024spiritlminterleavedspoken})        & 7B   & SQ, SA          & 0.0 & 2.0 \\
        Moshi(\citeyear{defossez2024moshispeechtextfoundationmodel})           & 7.6B & SQ, SA          & 0.0 & 2.8 \\
        Mini-Omni(\citeyear{DBLP:journals/corr/abs-2408-16725})       & 0.5B & TQ, SA          & 0.0 & 8.2 \\
        SpiritLM        & 7B   & TQ, SA          & 0.0 & 3.9 \\
        GLM-4-Voice(\citeyear{DBLP:journals/corr/abs-2412-02612})     & 9B   & TQ, SA          & \underline{0.1} & \underline{14.9} \\
        SpeechGPT(\citeyear{DBLP:conf/emnlp/ZhangLZZWZQ23})       & 7B   & SQ, TQ, TA, SA  & 0.0 & 6.6 \\
        \midrule
        Best Baseline            & 0.5B & TQ, TA, SA & \textbf{12.0} & \textbf{23.8} \\
        \bottomrule
    \end{tabular}
    \caption{Comparison of SLM baselines with various data formats on RoleTriviaQA ID test set, where TQ, TA, SQ, SA means text question, text answer, speech question, speech answer respectively.}
    \label{tab:role_play_slm}
\end{table}

\label{app:metrics_qa}
For SLM performance evaluation, we adopted:
\begin{itemize}
    \item \textbf{Success Rate (SR)} measures the model’s ability to learn token sequence patterns. It is defined as the ratio of successfully synthesized speech samples to the total number of test samples:
\[
\text{SR} = \frac{N_{\mathrm{success}}}{N_{\mathrm{total}}}.
\]
    \item \textbf{Word Error Rate (WER)} measures the accuracy of the semantic information of the generated speech. We employed the ASR model Whisper-large-v3\footnote{\scriptsize{\url{https://huggingface.co/openai/whisper-large-v3}}} to transcribe generated speech into text and then calculate the WER between the transcribed text and the text response.
    \item \textbf{UTMOS} assesses the acoustic quality of the generated speech. 
    We utilized the model provided by SaruLab\footnote{\scriptsize{\url{https://huggingface.co/spaces/sarulab-speech/UTMOS-demo/tree/main}}} to evaluate it.
    \item \textbf{Speaker Similarity (SIM)} reflects how similar it is between the generated speech and the reference speech. We considered the WavLM-TDCNN~\cite{DBLP:journals/jstsp/ChenWCWLCLKYXWZ22} speaker embedding model to assess speaker similarity.
\end{itemize}
2 more metrics are considered for the evaluation of the VQ reconstruction (See Appendix~\ref{app:codec_vq_append}):
\begin{itemize}
    \item \textbf{PESQ} quantifies the perceptual quality of speech signals by simulating human auditory perception under various distortions and degradations.
    \item \textbf{STOI} predicts speech intelligibility by analyzing time-frequency correlations between clean and processed speech signals.
\end{itemize}
For text quality evaluation, we utilized:
\begin{itemize}
    \item \textbf{Exact Match (EM)} is the percentage of generated answers that exactly match the ground truth, which is given by:
        \begin{equation}
        \text{EM} = \frac{1}{N} \sum_{i=1}^{N} \mathbf{1}(\hat{y_i} = y_i)
        \notag
        \end{equation}
    where \(N\) is the number of samples, \(\hat{y_i}\) is the generated answer, and \(y_i\) is the ground truth, with $\mathbf{1}$ representing the indicator function.
    
    \item \textbf{F1 score} represents the harmonic mean of precision and recall, formulated as:
    \begin{equation}
    \begin{aligned}
    \text{Precision} &= \frac{\text{TP}}{\text{TP} + \text{FP}} \\
    \text{Recall} &= \frac{\text{TP}}{\text{TP} + \text{FN}} \\
    \text{F1} &= \frac{2 \cdot \text{Precision} \cdot \text{Recall}}{\text{Precision} + \text{Recall}}
    \end{aligned}
     \notag
    \end{equation}
    where \textbf{TP} denotes words appearing in both ground truth and predicted answers, \textbf{FP} those exclusively in predictions, and \textbf{FN} those solely in ground truth.
\end{itemize}

We also utilized \textbf{Riemannian Distance} to quantify speech-text alignment. 

Riemann Distance captures the intrinsic geometry of the embedding space by modeling it as a curved manifold. The distance between the dominant speech hidden states ${H_{\text{\faFileSoundO}}}$ and text hidden states ${H_{\text{\faFileTextO}}}$ is computed as:
\begin{align}
    \textrm{Dist}(H_{\text{\faFileSoundO}}, H_{\text{\faFileTextO}}) = \sqrt{\sum_{1\le i\le d} \log^2(\lambda_i)} + ||\bm{\mu}_\text{\faFileSoundO} - \bm{\mu}_{\text{\faFileTextO}}||^{2},\nonumber
\end{align}
where $\lambda_i$ is the $i$-th positive real eigenvalue of $K_{{\text{\faFileTextO}}}^{-1} K_{{\text{\faFileSoundO}}}$, $\bm{\mu}_{\text{\faFileSoundO}}$ and  $\bm{\mu}_{\text{\faFileTextO}}$ are the centroids of speech and text embeddings respectively.
Here $K_{\text{\faFileSoundO}} \in \mathbb{R}^{d \times d}$ can be calculated from the SVD of the right singular matrices $V_{\text{\faFileSoundO}}$:
$
    K_{{\text{\faFileSoundO}}} = \frac{1}{n-1} V_{{\text{\faFileSoundO}}} \Sigma_{{\text{\faFileSoundO}}}^2 V_{{\text{\faFileSoundO}}}^\top.
$

\section{RoleTriviaQA Results on SLM Baselines}

\begin{table}[ht!]
    \centering
    \small
    \setlength{\tabcolsep}{1.5mm} 
    \begin{tabular}{ll|ccc}
    \toprule
    Fusion & Speech Head & UTMOS$\uparrow$ &WER$\downarrow$ & SIM$\uparrow$ \\ 
    \midrule
    Linear   &   Linear       & 3.75 & 9.42 & 0.45 \\
    Linear   &   MLP      & \underline{3.78} & \underline{7.28} & \underline{0.46} \\ 
    Linear   &   Transformers      & 3.31 & 22.19 & 0.41 \\ 
    MLP   &   MLP      & 3.75 & 11.8 & 0.45 \\ 
    MLP   &   Transformers      & 3.34 & 16.84 & 0.41 \\ 
    MLP   &   Linear      & \bf3.90 & \bf6.35 & \bf0.47 \\ 
    \bottomrule
    \end{tabular}
    \caption{Ablation analysis of decoupled speech representation integration in MTP, revealing the impact of fusion modules (Linear vs. MLP) and speech head architectures (Linear/MLP/Transformer).}
    \label{tab:mtp_selection}
\end{table}

Results in~\Cref{tab:role_play_slm} show that the best baseline (SLM trained with decoupled tokenizer FACodec) in our study reaches remarkably higher EM and F1 (12.0 and 23.8) than other SLM baselines, most of which fail to generate exactly correct answers (EM=0). This result confirms the superiority of decoupled SLMs to other baselines with different architectures.

\label{app:role_play_slm}

\section{Ablation Study}

\subsection{Effect of MTP Architecture}

\begin{table}[htbp]
    \centering
    \small
    \setlength{\tabcolsep}{1.5mm}
    \begin{tabular}{cc|cccc}
    \toprule
    Nq & {\begin{tabular}[c]{@{}c@{}}Data \\ Form\end{tabular}} & {\begin{tabular}[c]{@{}c@{}}Success Rate \\ (SR) $\uparrow$\end{tabular}} & UTMOS $\uparrow$ & WER $\downarrow$ & SIM $\uparrow$ \\ \midrule
    
    3 & FWI & \bf 0.87 & 3.93 & \bf 6.07 & \bf 0.49 \\
    6 & FWI  & 0.82 & \bf 4.00 & 8.23 & 0.34 \\
    \bottomrule
    \end{tabular}
    \caption{Results of the ablation analysis of the number of codebook layers (Nq) used in FACodec of SLM's NTP training.}
    \label{tab:ntp_selection_layer}
\end{table}
\label{app:mtp_selection}
\Cref{tab:mtp_selection} compares different MTP architectures (Fusion and Speech Head).
First, controlling the speech head, compared to Linear, the MLP fusion improves performance in most cases (2/3). The results demonstrate the importance of non-linear fusion to decouple speech features.
Transformer-based heads result in substantial degradation in the WER, regardless of the fusion method, likely due to overfitting or incompatibility with the fused representations. 
The MLP–Linear configuration achieves the strongest overall performance in all metrics. Therefore, we select it as the default experimental setting for MTP.

\subsection{Effect of Number of Codebook Layers in SLMs}
\label{app:ntp_selection_layer}
\begin{table}[htbp]
    \centering
    \small
    \setlength{\tabcolsep}{1.5mm}
    \begin{tabular}{cc|cccc}
    \toprule
    Nq & {\begin{tabular}[c]{@{}c@{}}Data \\ Form\end{tabular}} & {\begin{tabular}[c]{@{}c@{}}Success Rate \\ (SR) $\uparrow$\end{tabular}} & UTMOS $\uparrow$ & WER $\downarrow$ & SIM $\uparrow$ \\ \midrule
    
    3 & CWI  & 0.80 & 3.30 & 22.21  & 0.43 \\
    3 & FWI & \bf 0.87 & \bf 3.93 & \bf 6.07 & \bf 0.49 \\

    \bottomrule
    \end{tabular}
    \caption{Results of ablation analysis of speech token organization pattern (Data Form) used in FACodec of SLM's NTP training.}
    \label{tab:ntp_selection_form}
\end{table}
As indicated in Table~\ref{tab:ntp_selection_layer}, the 3-layer setting (1 prosody layer and 2 content layers) in FACodec-NTP training exhibits superior overall performance and a higher success rate compared to the full 6-layer (adding 3 acoustic layers based on the previous setting) configuration. 
Therefore, we chose the 3-layer setting for the main NTP and MTP experiments.

\subsection{Effect of Speech Token Organization Pattern}

\label{app:ntp_selection_form}
We compared two speech token organization patterns: Frame-Wise Interleaving (FWI) and Chunk-Wise Interleaving (CWI).
\paragraph{FWI} organizes tokens as $[p_i, c_{i,1}, c_{i,2}]$ per frame, where $p_i$ denotes prosody tokens and $c_{i,j}$ represents content tokens for the $i$-th frame.

\paragraph{CWI} groups tokens into chunks $[p_{i:i+l}, c_{i:i+l,1}, c_{i:i+l,2}]$ with layer length $l$ (typically 80), where the final chunk may have $\leq$80 tokens.

~\Cref{tab:ntp_selection_form} shows that FWI outperforms CWI on all metrics, especially on WER (6.07 vs. 22.21), 
which means the CWI‐trained model aligns text and speech modalities less effectively than its FWI‐trained counterpart. Therefore, we opt for FWI as the default setting for our main experiments.
\subsection{Effect of RoleTriviaQA data format}
\begin{table*}[htb]
\centering
\small
\begin{tabular}{l|l|cccccc}
\toprule
\multirow{2}{*}{\textbf{Data Format}} & \multirow{2}{*}{\textbf{Model}} & \multicolumn{3}{c}{\textbf{ID}} & \multicolumn{3}{c}{\textbf{OOD}} \\
\cmidrule(lr){3-5} \cmidrule(lr){6-8}
 & & EM$\uparrow$ & F1$\uparrow$ & SIM$\uparrow$ & EM & F1 & SIM \\
\midrule

\multirow{3}{*}{\{SQ, SA\}} 
 & StableCodec-NTP-1H & 0.0* & 0.6* & 0.21 & 0.1* & 0.2* & 0.21 \\
 & FACodec-NTP-1H     & 0.5* & 2.4* & 0.60 & 0.4* & 1.0* & 0.60 \\
 & FACodec-MTP-12H    & 0.6* & 2.5* & 0.60 & 0.3* & 1.6* & 0.59 \\
\midrule

\multirow{3}{*}{\{TQ, SA\}} 
 & StableCodec-NTP-1H & 0.0* & 0.3* & 0.17 & 0.0* & 0.2* & 0.16 \\
 & FACodec-NTP-1H     & 0.5* & 2.4* & 0.60 & 0.8* & 1.6* & 0.60 \\
 & FACodec-MTP-12H    & 0.7* & 3.1* & 0.59 & 0.3* & 1.6* & 0.59 \\
\midrule

\multirow{3}{*}{\{TQ, TA, SA\}} 
 & StableCodec-NTP-1H & 7.5 & 13.8 & 0.17 & 0.6 & 3.6 & 0.16 \\
 & FACodec-NTP-1H     & \underline{12.0} & \underline{23.8} & \underline{0.61} & 9.0 & 17.9 & \textbf{0.61} \\
 & FACodec-MTP-12H    & \textbf{18.0} & \textbf{29.6} & \textbf{0.61} & \textbf{10.8} & \textbf{21.4} & \underline{0.60} \\
\bottomrule
\end{tabular}
\caption{Comparison of various data format of RoleTriviaQA on different tokenizers.\ * indicates the text answers are obtained by ASR.}
\label{tab:data_format_ablation}
\end{table*}
\label{app:data_format_ablation}

We evaluate three data formats for RoleTriviaQA:

\begin{itemize}
    \item ${\{X_{\text{\small \faUser}}, X^{(q)}_{\text{\small\faFileSoundO}}, X^{(a)}_{\text{\small\faFileSoundO}}}\}$ (i.e., \{SQ, SA\})

    \item ${\{X_{\text{\small \faUser}}, X^{(q)}_{\text{\small\faFileTextO}}, X^{(a)}_{\text{\small\faFileSoundO}}}\}$ (i.e., \{TQ, SA\})

    \item $\{{X_{\text{\small \faUser}}, X^{(q)}_{\text{\small\faFileTextO}}, X^{(a)}_{\text{\small\faFileTextO}}, X^{(a)}_{\text{\small\faFileSoundO}}}\}$ (i.e., \{TQ, TA, SA\})
\end{itemize}

\Cref{tab:data_format_ablation} compares these formats for SLMs trained with StableCodec and FACodec under both NTP and MTP settings\footnote{Speaker embedding $X_{\text{\small \faUser}}$ is unavailable for coupled tokenizers.}.

Results show that the {TQ, TA, SA} format performs best, achieving top scores (EM 18.0, F1 29.6) with FACodec and MTP heads. It also significantly improves performance for StableCodec-trained SLMs compared to other formats (EM 7.5 vs. 0.0; F1 13.8 vs. 0.6).

\begin{figure*}[t]
    \centering
    \begin{subfigure}{0.32\textwidth}
        \centering
        \includegraphics[width=\textwidth]{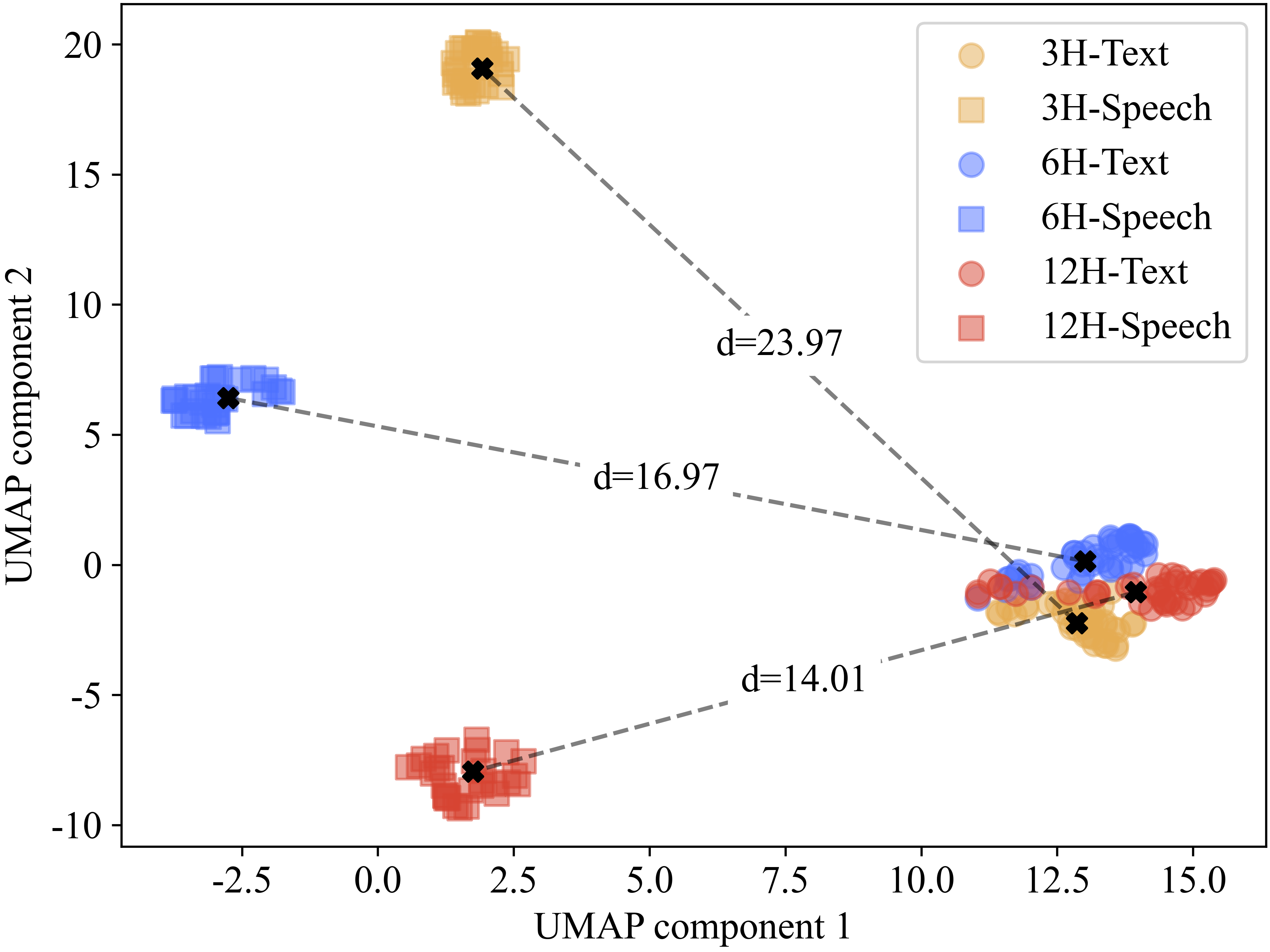}
        \caption{UMAP of word embeddings}
        \label{fig:word_emb}
    \end{subfigure}
    \hfill
    \begin{subfigure}{0.32\textwidth}
        \centering
        \includegraphics[width=\textwidth]{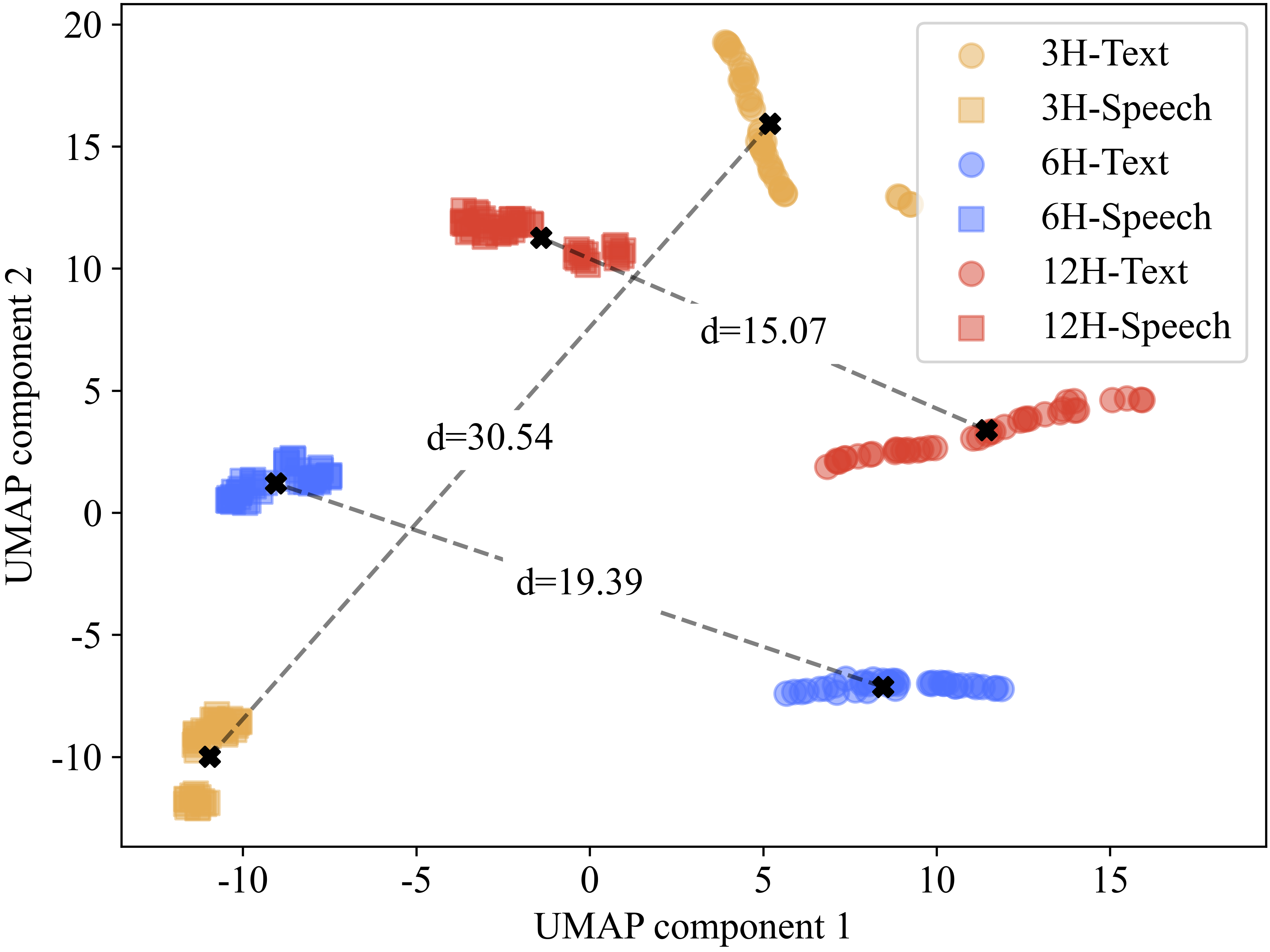}
        \caption{UMAP of middle layer}
        \label{fig:mid_layer}
    \end{subfigure}
    \hfill
    \begin{subfigure}{0.32\textwidth}
        \centering
        \includegraphics[width=\textwidth]{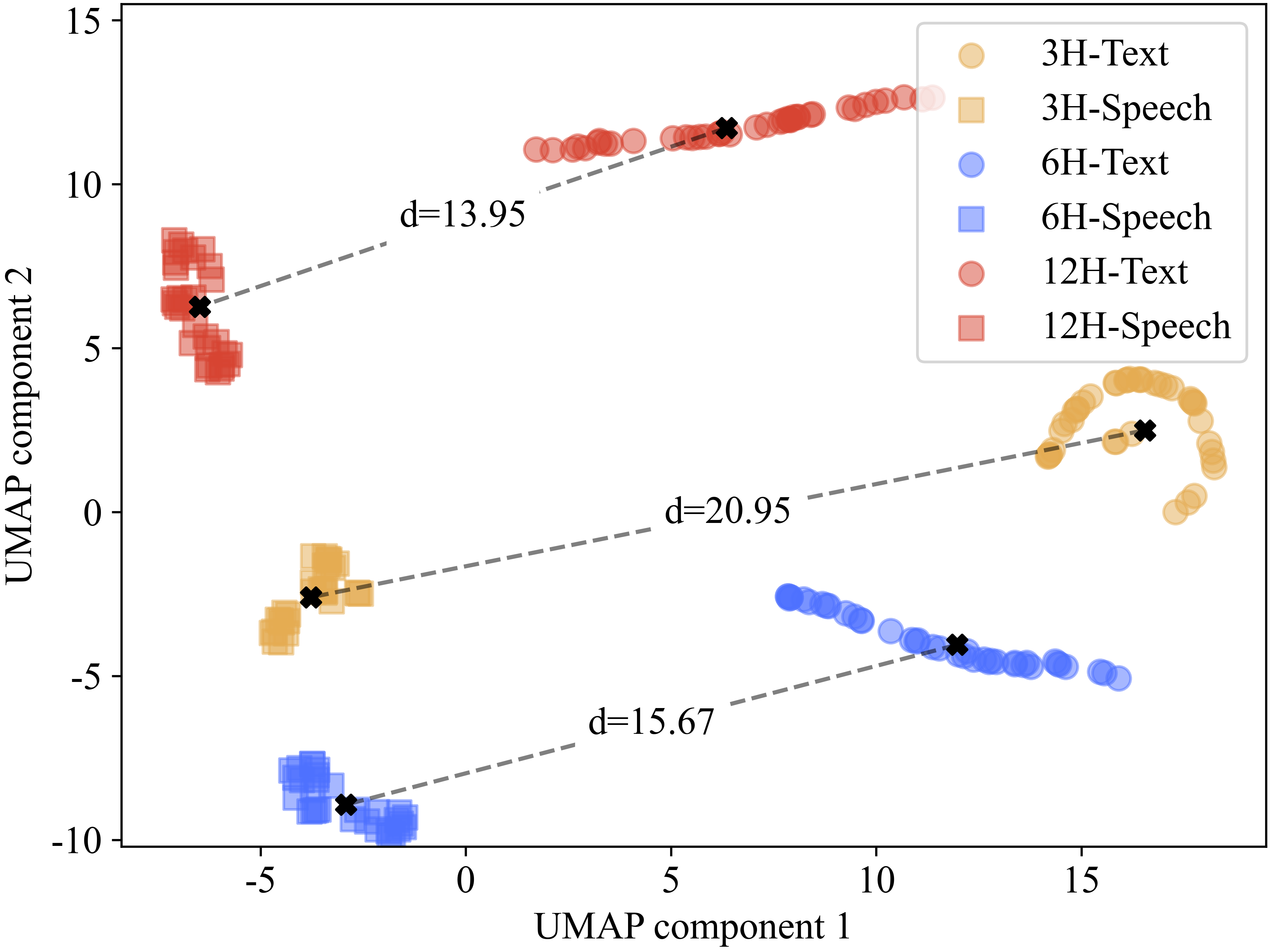}
        \caption{UMAP of last layer}
        \label{fig:last_layer}
    \end{subfigure}
    
    \caption{UMAP visualization of speech and text embedding distributions across the word embeddings, the middle and the last layer in MTP models of different numbers of speech heads (3H, 6H, and 12H), illustrating how the relative distance between modalities changes when the number of speech heads increases.} 
    \label{fig:speech_text_umap}
\end{figure*}

\section{Analysis}
\subsection{Quantitative Analysis of MTP’s Impact on Cross-Modal Alignment}
\label{app:umap_quantitative}
~\Cref{fig:speech_text_umap} displays UMAP projections of speech and text representations from different layers in SLM. These projections illustrate the impact of different compression rates on the SLM's training, revealing three key observations.

First, speech and text representations consistently form distinct modality-specific clusters across all layers and compression rates. 
Second, all models show a layer-wise refinement. Although cross-modal distances increase in intermediate layers, deeper layers progressively reduce the gap between modalities, where cross-modal distances are lower in the final layer compared to the word embedding layer.
Third, higher compression consistently reduces cross-modal distance. Specifically, 12H shows the closest alignment in both word embedding ($14.01$), mid layers ($15.07$), and last layers ($13.95$), followed by 6H and 3H, indicating that compression facilitates tighter alignment between modalities.

\subsection{Effect of Tokenizer-MTP Combination}
\label{app:mtp_role}

\begin{table}[ht!]
    \centering
    \small
    \setlength{\tabcolsep}{1.8mm} 
    \begin{tabular}{lc|ccc}
    \toprule
    Models &  CR  & UTMOS$\uparrow$ & WER$\downarrow$  & SIM$\uparrow$ \\ 
    \midrule
    \multicolumn{5}{c}{\textit{Coupled Tokenizers trained with SLM}} \\ 
    \midrule
    BigCodec   &   1      & \underline{3.93} & 13.63 & - \\ 
    BigCodec   &   2      & \bf3.97 & 15.48 & - \\ 
    BigCodec   &   4      & 3.29 & 13.48 & - \\ 
    BigCodec   &   8      & 1.94 & 46.82 & - \\ 
    \midrule
    \multicolumn{5}{c}{\textit{Semi-Decoupled Tokenizers trained with SLM}} \\ 
    \midrule
     SpeechTokenizer &   1      & 3.86 & 13.13 & - \\ 
    SpeechTokenizer &   5      & 3.57 & 12.9 & - \\
     
    SpeechTokenizer &   10      & 3.23 & 9.67 & - \\
    \midrule
    \multicolumn{5}{c}{\textit{Decoupled Tokenizers trained with SLM}} \\ 
    \midrule
    FACodec   &   1      & 3.93 & 6.07 & \bf{0.49} \\ 
    FACodec   &   3      & 3.90 & 6.35 & \underline{0.47} \\ 
    FACodec   &   6     & 3.75 & \underline{4.36} & 0.47 \\ 
    FACodec   &   12      & 3.67 & \bf 3.01 & 0.47 \\ 
    \bottomrule
    \end{tabular}
    \caption{Performance comparison of coupled (BigCodec), semi-decoupled (SpeechTokenizer) and decoupled (FACodec) tokenizers trained with SLM at different Compression Ratios (CR).}\label{tab:mtp_couple_vs_decoupe}
\end{table}

Table~\ref{tab:mtp_couple_vs_decoupe} investigates the impact of different tokenizers trained on MTP-based SLMs.
We select semi-decoupled (SpeechTokenizer), decoupled (FACodec) tokenizers for comparison, and choose BigCodec as our coupled tokenizer baseline, due to its 80 TPS for fair analysis and strong performance in coupled tokenizers as shown in \Cref{tab:codec_compare}.

Primarily, coupled models experience severe degradation as compression increases: UTMOS drops sharply from 3.97 to 1.94, and WER rises to 46.82 at 8× compression.
Additionally, semi-decoupled models show some WER improvement but with a slight decline in UTMOS, even with a $10\times$ compression rate.
Finally, the decoupled architecture maintains speaker similarity and consistently reduces the WER under higher compression, with only a minor reduction in audio quality. Remarkably, even at a $12\times$ compression, the WER is close to the ground truth.
Our above results indicate that MTP architectures achieve the best performance with decoupled tokenizers, followed by semi-decoupled and coupled tokenizers.

\section{Details of RoleTriviaQA}
\label{app:role_play_dataset}

\begin{table*}[htb!]
\centering
\small
\setlength{\tabcolsep}{2mm}
\begin{tabular}{cc|c|c|c}
\toprule
\multirow{2}{*}{Speaker} & \multirow{2}{*}{Gender}          & \multirow{2}{*}{Train Dataset Size}  & \multicolumn{2}{c}{Evaluation Dataset Size} \\ \cmidrule(lr){4-5}
                      &               &                     & \textbf{TriviaQA (ID)}  &\textbf{Web Questions (OOD)} \\ \midrule
\rowcolor{gray!10} \multicolumn{5}{c}{\textbf{Seen Roles}}                      \\
Madame Ping           & Old Female    & 14,244              & 100 & 65          \\
Venti                 & Teenage Male  & 13,664              & 100 & 47          \\
Xiao                  & Male          & 14,111              & 100 & 54          \\
Kaedehara Kazuha      & Male          & 13,570              & 100 & 55          \\
Arataki Itto          & Male          & 13,751              & 100 & 58          \\
Kamizato Ayato        & Male          & 14,025              & 100 & 64          \\
Furina                & Female        & 14,266              & 100 & 58          \\
Nahida                & Female        & 13,357              & 100 & 69          \\
Yae Miko              & Female        & 13,807              & 100 & 73          \\
Keqing                & Female        & 13,589              & 100 & 80          \\
\midrule
\rowcolor{gray!10} \multicolumn{5}{c}{\textbf{Unseen Roles}}                    \\
Albedo                & Male          & -                  & 100 & 45           \\
Wriothesley           & Male          & -                  & 100 & 70           \\
Jean                  & Female        & -                  & 100 & 64           \\
Lisa                  & Female        & -                  & 100 & 62           \\
Eula                  & Female        & -                  & 100 & 62           \\
\midrule
\multicolumn{2}{c|}{\textbf{Total}}   & 138,384            & 1,500 & 926          \\
\bottomrule
\end{tabular}
\caption{Summary of the roles and the corresponding train/test dataset size of RoleTriviaQA.}
\label{tab:qa_data_summary}
\end{table*}

\subsection{Data Construction} 
\paragraph{Data Composition}
\Cref{tab:qa_data_summary} demonstrates the detailed composition of RoleTriviaQA along with the selected speakers. The ID evaluation samples are collected from the TriviaQA test set, and OOD samples are gathered from the Web Questions subset of OpenAudioBench. Notably, we added about 26,000 text QA training samples filtered from LLaVA-Instruct-150k\footnote{\scriptsize{https://huggingface.co/datasets/liuhaotian/LLaVA-Instruct-150K}} during training to preserve the original textual ability of the SLM.

\paragraph{Role-Playing Speaker Selection}
For \textit{seen} speakers, we selected 10 representative speakers, including 5 males and 5 females, from 130 common speakers based on the character's hotness ranking and the quality of synthesized speeches while maintaining age diversity.
For \textit{unseen} speakers, we selected 5 speakers, including 2 males and 3 females, from the remaining 120 speakers with the same criteria.
\paragraph{Speech Synthesis}
For each character's speech synthesis, we select high-quality reference speeches by combining speaker embedding similarity metrics with manual filtering as a comprehensive criterion. Then we leverage the CosyVoice 2
\footnote{\scriptsize{\url{https://huggingface.co/FunAudioLLM/CosyVoice2-0.5B}}} 
system to generate role-playing speech from text answers. Finally, all speeches are resampled to 16 kHz to fit the tokenization requirement.

\subsection{Samples of RoleTriviaQA}
\label{app:role_play_dataset_sample}
\tcbset{
  mybox/.style={
    enhanced,
    colback=gray!5,
    colframe=gray!80!black,
    fonttitle=\bfseries,
    arc=5pt,                    
    rounded corners,            
    boxrule=0.8pt,
    fontupper=\small,
    top=1mm, bottom=1mm, left=1mm, right=1mm
  }
}
\begin{minipage}[htb]{0.48\textwidth}
\begin{tcolorbox}[mybox]
\textit{Example 1} \\[3pt]
\textbf{Text Query:} Who found the remains of the Titanic? \\
\textbf{Text Answer:} The answer is \textit{robert duane ballard} \\[2pt]
\textbf{Speech Answer Tokens\footnote{\scriptsize{\texttt{<a*>} are prosody tokens, \texttt{<b*>} and \texttt{<c*>} are content tokens.}}:} \\
\texttt{\small <a106><b515><c195><a106><b515>\\
<c850><a308><b515><c578><a308>\\
<b515><c53><a335><b515><c970>}... (more omitted) \\[2pt]
\textbf{Speaker:} Madame Ping
\\
\\
\textit{Example 2} \\[3pt]
\textbf{Text Query:} What do internet retailers call the Monday after Black Friday? \\[2pt]
\textbf{Text Answer:} The answer is \textit{cybermonday} \\[2pt]
\textbf{Speech Answer Tokens:} \\
\texttt{\small <a482><b261><c101><a566><b363>\\
<c983><a846><b247><c590><a111>\\
<b56><c983><a354><b263><c921>}... (more omitted) \\[2pt]
\textbf{Speaker:} Nahida
\end{tcolorbox}
\end{minipage}
\hfill

\subsection{Prompt template used in Knowledge-Role Joint Fine-Tuning}
\label{app:prompt_templates}
\begin{tcolorbox}[
    colback=gray!5,
    arc=5pt,                    
    rounded corners,    
    colframe=gray!80!black,
    fonttitle=\bfseries,
    boxrule=0.8pt,
    top=3mm,
    bottom=3mm,
    left=2mm,
    right=2mm,
    beforeafter skip=5mm]

\textbf{\small Prompt:}
\small
You are a helpful, harmless, and honest AI assistant. You are capable of handling speech and text inputs while responding to the user with accurate, polite, and safe cross-modal outputs. \textbf{[Human]}: Respond to this text question. You can first generate a text answer and read this answer. You should speak like: \textit{\{spk-emb-placeholder\}}. Question: \{\textit{query}\} \textbf{[Assistant]}: \textbf{[ta]}\{\textit{text-response}\}\textbf{[ua]}\{\textit{speech-response}\}.
\end{tcolorbox}

\section{Multi Token Prediction}
\label{app:related_word_mtp}

Multi Token Prediction (MTP) is leveraged in LLMs to accelerate inference through three principal strategies: 1) Multi-head architectures enhance generation quality via parallel output heads~\cite{DBLP:conf/icml/GloeckleIRLS24}; 2) Tree-based systems like MEDUSA~\cite{DBLP:conf/icml/CaiLGPLCD24} refine candidate selection through hierarchical validation; 3) Scalable frameworks such as DeepSeek-V3~\cite{deepseekai2025deepseekv3technicalreport} enable efficient large-scale training. Recent adaptations extend MTP to speech processing: SLAM-Omni~\cite{chen2024slamomnitimbrecontrollablevoiceinteraction} employs audio token grouping for computational efficiency but neglects paralinguistic modeling through semantic-focused processing, while VocalNet \cite{wang2025vocalnetspeechllmmultitoken} achieves localized acceleration via DeepSeek-V3-style parallel prediction yet faces pipeline complexity from flow-matching requirements.

\end{document}